\def\paperTitle{\text{TeG-DG}: Textually Guided Domain Generalization for Face Anti-Spoofing}

\def\authorBlock{
    Lianrui Mu\thanks{Equal contribution} \qquad
    Jianhong Bai\footnotemark[1] \qquad
    Xiaoxuan He \qquad
    Jiangnan Ye \\
    Xiaoyu Liang\qquad
    Yuchen Yang \qquad
    Jiedong Zhuang \qquad
    Haoji Hu\thanks{Corresponding Author}\\
    Zhejiang University \\
    {\tt\small mulianrui@zju.edu.cn}
}

\newif\ifreview 
\newif\ifarxiv \newcommand{\arxiv}{\arxivtrue}
\newif\ifcamera 
\newif\ifrebuttal 

\arxiv % \review OR \arxiv OR \cameraready

\pdfoutput=1
\documentclass[10pt,twocolumn,letterpaper]{article}

\ifreview \usepackage[review]{cvpr} \fi
\ifarxiv \usepackage[pagenumbers]{cvpr} \fi
\ifrebuttal \usepackage[rebuttal]{cvpr} \fi
\ifcamera \usepackage{cvpr} \fi

\usepackage{graphicx}	
\usepackage{amsmath}	
\usepackage{amssymb}	
\usepackage{booktabs}
\usepackage{times}
\usepackage{microtype}
\usepackage{epsfig}
\usepackage[table,xcdraw,dvipsnames]{xcolor}
\usepackage{caption}
\usepackage{float}
\usepackage{placeins}
\usepackage{color, colortbl}
\usepackage{stfloats}
\usepackage{enumitem}
\usepackage{tabularx}
\usepackage{xstring}
\usepackage{multirow}
\usepackage{xspace}
\usepackage{url}
\usepackage{subcaption}
\usepackage{xcolor}
\usepackage[hang,flushmargin]{footmisc}

\usepackage{algorithm}
\usepackage{algpseudocode}
\usepackage{xcolor}
\usepackage{tabularx}
\usepackage{fontawesome}

\definecolor{graycolor}{gray}{0.9}

\ifcamera \usepackage[accsupp]{axessibility} \fi

% ...

\ifarxiv  \fi

\newcommand{\R}[1]{{%
    \textbf{%
        \ifstrequal{#1}{1}{\textcolor{red}{R#1}}{%
        \ifstrequal{#1}{2}{\textcolor{blue}{R#1}}{%
        \ifstrequal{#1}{3}{\textcolor{magenta}{R#1}}{%
        \ifstrequal{#1}{4}{\textcolor{teal}{R#1}}{%
                           \textcolor{cyan}{R#1}%
        }}}}%
    }%
}}

\usepackage{xr-hyper}

\makeatletter
\newcommand*{\addFileDependency}[1]{
  \typeout{(#1)}
  \@addtofilelist{#1}
  \IfFileExists{#1}{}{\typeout{No file #1.}}
}

\makeatother

\definecolor{cvprblue}{rgb}{0.21,0.49,0.74}
\usepackage[pagebackref,breaklinks,colorlinks,citecolor=cvprblue]{hyperref}
\usepackage[capitalize]{cleveref}
\crefname{section}{Sec.}{Secs.}
\crefname{table}{Table}{Tables}
\crefname{figure}{Fig.}{Figs.}

\usepackage{indentfirst}
\begin{document}
%% TITLE
\title{\paperTitle}
\author{\authorBlock}
\maketitle

\begin{abstract}
Enhancing the domain generalization performance of Face Anti-Spoofing (FAS) techniques has emerged as a research focus.
Existing methods are dedicated to extracting domain-invariant features from various training domains. 
Despite the promising performance, the extracted features inevitably contain residual style feature bias (\emph{e.g.}, illumination, capture device), resulting in inferior generalization performance. 
In this paper, we propose an alternative and effective solution, the \textbf{Te}xtually \textbf{G}uided \textbf{D}omain \textbf{G}eneralization (\texttt{TeG-DG}) framework, which can effectively leverage text information for cross-domain alignment. 
Our core insight is that text, as a more abstract and universal form of expression, can capture the commonalities and essential characteristics across various attacks, bridging the gap between different image domains. 
Contrary to existing vision-language models, the proposed framework is elaborately designed to enhance the domain generalization ability of the FAS task. 
Concretely, we first design a Hierarchical Attention Fusion (HAF) module to enable adaptive aggregation of visual features at different levels; Then, a Textual-Enhanced Visual Discriminator (TEVD) is proposed for not only better alignment between the two modalities but also to regularize the classifier with unbiased text features.
\texttt{TeG-DG} significantly outperforms previous approaches, especially in situations with extremely limited source domain data ($\sim$\textbf{14\%} and $\sim$\textbf{12\%} improvements on HTER and AUC respectively), showcasing impressive few-shot performance.
The code will be publicly available.
\end{abstract}
\section{Introduction}
\label{sec:intro}

\begin{figure}[ht]
\centering
\includegraphics[width=3in]{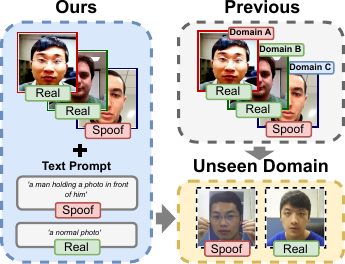}
\vspace{-0.2cm}
\caption{\textbf{Comparison with previous Face Anti-Spoofing (FAS) methods.} Our approach leverages the text information for better generalization \textit{without} using domain labels.}
\vspace{-0.5cm}
\label{fig_showcase}
\end{figure}

Facial recognition technology~\cite{taigman2014deepface, schroff2015facenet, deng2019arcface} has become a pivotal component in various applications, including mobile device authentication~\cite{fathy2015face}, electronic payments~\cite{rija2022payment}, and cybersecurity~\cite{slusky2020cybersecurity}. However, the integrity of this technology is constantly under threat from various presentation deception, \textit{e.g.} printed photo~\cite{peixoto2011face, tan2010face}, mask~\cite{kose2013shape, manjani2017detecting, yu2020fas, liu2022contrastive}, video replay \cite{pinto2015using, boulkenafet2017oulu, chingovska2012effectiveness, wen2015face, zhang2012face, zhang2020celeba}, \textit{etc}. To counter these attacks, a spectrum of Face Anti-Spoofing (FAS) works~\cite{pan2007eyeblink, lbp} has been proposed, leveraging either handcrafted~\cite{de2013lbp, patel2016secure, hog, rppg, komulainen2013context, yang2013face, boulkenafet2015face}, or deep learning features~\cite{feng2016integration, wang2022learning, yang2014learn, yu2020searching, yu2021dual, yu2021revisiting} for spoof detecting. While these approaches demonstrate commendable test performance under intra-dataset scenarios, their generalization capability is severely compromised when confronted with unseen domains~\cite{yu2022deep, shao2019multi}.

To this end, recent FAS studies integrate ideas from the field of domain adaptation (DA)~\cite{li2018unsupervised, zhou2020domain, wang2021self, zhou2022generative} and domain generalization (DG)~\cite{SSDG_FAS, liu2021adaptive, liu2021dual, zhou2022adaptive, SSAN_FAS, IADG_FAS}. Specifically, DA~\cite{long2015learning, long2017deep, wang2018deep} aims to narrow the gap between the source and target domains by fine-tuning models with target domain samples, requiring the collection of target domain data. While DG~\cite{blanchard2011generalizing, huang2017arbitrary, zhou2022domain} aims to learn universally applicable feature spaces, ensuring a consistent representation of similar categories across various domains. In general, both DA and DG encourage the model in learning \textit{domain-invariant} features across multiple training domains. Despite their promising results, DA-based methods~\cite{wang2019improving, jia2021unified, quan2021progressive} still rely on a modest amount of target domain images for adaptation, which is not always available in practice~\cite{yu2022deep}. On the other hand, current DG-based methods~\cite{shao2019multi, chen2021generalizable} are not supervised to learn robust features that can transfer to broader unseen domains, causing them to be deceived by the style features inherent to the limited training domains~\cite{SSDG_FAS, SSAN_FAS}.

In this paper, we proposed a novel and principal method to exploit the textual information to improve the generalization ability of FAS. As illustrated in Fig.~\ref{fig_showcase}, we supplement the single-modality visual features with text descriptions that match different types of attacks or real cases for better
generalization. Our core insight is that text descriptions possess a natural universality across different domains and can be leveraged to bridge the gap between various visual domains (\textit{e.g.}, images captured by different devices or having distinct backgrounds). However, we empirically discover that a naive migration of existing vision-language models~\cite{CLIP, BLIP, kirillov2023segment} to FAS results in inferior performance (see Sec. \ref{sec:few_shot} and Appendix). To this end, we devise a new framework, Textually Guided Domain Generalization (\texttt{TeG-DG}), to better exploit textual information for both visual feature learning and decision boundary optimization. 

The pipeline of the proposed framework is illustrated in Fig.~\ref{fig_pipeline}. \texttt{TeG-DG} comprises a Text Prompter (TP) to synthesize paired image-text data for training. During training, a Hierarchical Attention Fusion (HAF) module and a Textual-Enhanced Visual Discriminator (TEVD) are designed to work collaboratively to tackle the domain generalization problem in FAS. Specifically, since different attacks may affect features at different granularities, the hierarchical attention fusion module aims to adaptively merge both the local texture features and the high-level semantics of an input image. On the other hand, the textual-enhanced visual discriminator incorporates textual supervision by matching the aggregated visual features with the corresponding text descriptions, which act as prototypes of each category, and regularizing the decision boundaries of the classifier. 
Experimental results on various datasets demonstrate the effectiveness of \texttt{TeG-DG}. We also conduct comprehensive analyses to understand the designed method. In summary, the main contributions of the paper are as follows:

\begin{itemize} 
\item \textit{New insight.} We propose to solve the domain generalization problem in face anti-spoofing by introducing domain-invariant supervision from textual description.

\item \textit{New framework.} We design a novel yet easy-to-use FAS framework, which is composed of a hierarchical attention fusion module and a textual-enhanced visual discriminator, both of which are elaborately designed to enhance the domain generalization ability of the FAS task.

\item \textit{New scenarios and compelling empirical results.} We conducted extensive experiments on widely used benchmark datasets to verify and understand the effectiveness of the proposed method. We further extend FAS to few-shot scenarios and scenarios with extremely limited source domain data, which are more common in practice.

\end{itemize}
%\vspace{-0.6cm}
\section{Related Work}
\label{sec:related}
%\vspace{-0.1cm}

\textbf{Face Anti-Spoofing}
Early attempts at Face Anti-Spoofing (FAS) predominantly utilized traditional hand-crafted features such as SIFT~\cite{patel2016secure}, LBP~\cite{lbp}, and HOG~\cite{hog}. They primarily use color spaces and temporal information to differentiate between real and spoofed faces. The advent of deep learning, particularly Convolutional Neural Networks (CNNs), marked significant advancements in FAS. This evolution included the integration of auxiliary signals like depth maps~\cite{atoum2017face, shao2019multi}, r-ppg signals~\cite{rppg} or reflection map~\cite{yu2020face} to enhance detection capabilities. Although these techniques show promising intra-dataset performance, they exhibit severe performance degradation when applied to target datasets due to the domain shifts.

Follow-up studies improve the ability of FAS methods on domain generalization. Apart from domain adaptation-based methods~\cite{li2018unsupervised, zhou2020domain, wang2021self, zhou2022generative}, meta-learning techniques~\cite{du2022energy, liu2021adaptive, liu2021dual, shao2020regularized, zhou2022adaptive} and domain generalization-based methods have become a promising direction in the field of FAS in recent years. Typical techniques in DG-based methods involve using adversarial learning to form compact distributions for real faces while distancing spoofed faces~\cite{SSDG_FAS, shao2019multi, SSAN_FAS, IADG_FAS}. In particular, MADDG~\cite{shao2019multi} and SSAN~\cite{SSAN_FAS} employed domain labels to generate assembled features to learn a domain-invariant feature space from the training domains. However, it was noted that the artificial domain labels used in these methods are coarse and do not accurately reflect the real domain distributions.
Consequently, IADG~\cite{IADG_FAS} attempted to alleviate this issue by using an instance feature whitening method without relying on domain labels. A recent work~\cite{liao2023domain} also utilizes Transformer models in FAS to achieve stronger representational capabilities. Additionally, FLIP ~\cite{Srivatsan_2023_ICCV} shows that direct finetuning of a multimodal pre-trained ViT to align with text prompts achieves better FAS generalizability. Compared with FLIP~\cite{Srivatsan_2023_ICCV}, we delve deeper into the impact of text prompts on generalization ability~\ref{appen: prompt_numbers}. We use the HAF module to generate superior visual representations and do not require the involvement of text prompts during inference.
Nevertheless, having access to multiple training domains is challenging in practice, which limits the application of existing methods. In this paper, we introduce a novel approach to address these issues by incorporating domain-invariant textual information into the FAS task.

\textbf{Vision-Language Models}
The recent development in visual-language model (VLM) research reveals the critical role of textual data in augmenting visual task performance. Foundational VLMs like CLIP~\cite{CLIP} and related works~\cite{li2021align, BLIP, furst2022cloob, li2021supervision} completed mapping images and text to a shared representation space, setting a platform in tasks for zero-shot learning and image-text classification. This confluence of modalities not only enriches the information base for these models but also addresses inherent challenges in image interpretation, such as ambiguity, noise, and the need for fine-grained recognition.
Building upon this groundwork, a few works illustrated the potential of deeply integrated textual information in VLMs. CMA-CLIP~\cite{liu2021cma} adeptly integrates sequence-wise and modality-wise attentions to synergize image and text data, thus facilitating enhanced multi-task classification. Menon \emph{et al.}~\cite{menon2022visual} leverage detailed descriptors from large language models, presenting an interpretable paradigm for image classification that significantly improves visual categorization. Lin \emph{et al.}~\cite{lin2023multimodality} propose a novel cross-modal adaptation method that utilizes cues from visual, textual, and audio modalities, effectively transcending the limitations of vision-only learning methods. CuPL~\cite{pratt2023does} demonstrates that textual descriptions can effectively guide models in recognizing and classifying unseen objects, even in the absence of extensive training datasets. These studies underscore the capability of pre-trained visual-language models to infuse textual supervision, thereby enabling the use of textual information for improved generalization beyond the training domains. We introduce an innovative Textual-Enhanced Visual Discriminator (TEVD) module in our work, which leverages text information for classification assistance and regularization. This allows for the organic integration of features provided by the language model into the FAS task.

\section{Textually-Guided Domain Generalization}

\begin{figure*}[t]
\centering
\includegraphics[width=6.5in]{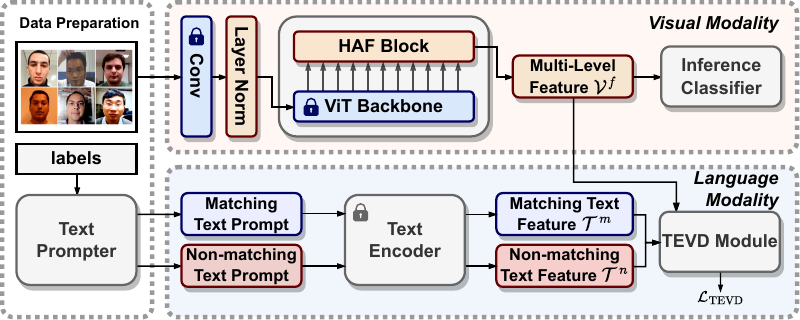}

\vspace{-0.1cm}

\caption{\textbf{Overview of the proposed Textually Guided Domain Generalization (\texttt{TeG-DG}) framework.} The framework contains Text Prompter (TP) for text prompt generation, the Hierarchical Attention Fusion(HAF) module for fused visual feature extraction, and the Texual-Enhanced Visual Discriminator (TEVD) for integrating text information.
}

\vspace{-0.4cm}
\label{fig_pipeline}
\end{figure*}

In this section, we will present a detailed description of the proposed Textually Guided Domain Generalization (\texttt{TeG-DG}) framework, as illustrated in Fig.~\ref{fig_pipeline}. In Sec.~\ref{sec:data_prepare}, we briefly describe the construction process of the image-text pair data for multi-modal training. In Sec. \ref{sec:method}, we introduce the two main components of \texttt{TeG-DG}, namely, the Hierarchical Attention Fusion (HAF) module (Sec.~\ref{sec:method_part1}) and the Texual-Enhanced Visual Discriminator (TEVD) (Sec.~\ref{sec:method_part2}). They work collaboratively to learn discriminative and domain-invariant features for better generalization.
\subsection{Data Preparation}
\label{sec:data_prepare}

\begin{figure}[t]
\centering
\includegraphics[width=3.25in]{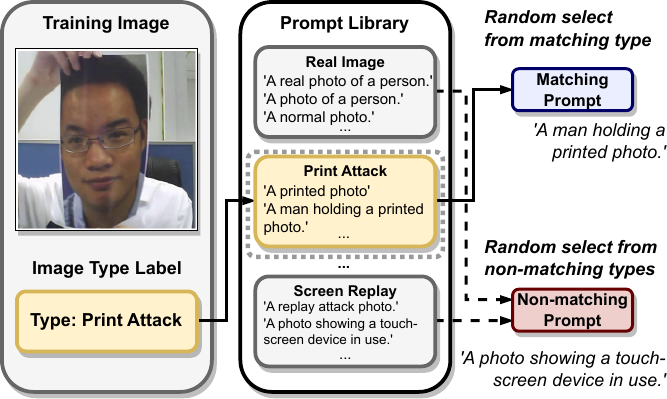}

\vspace{-0.15cm}
\caption{\textbf{Text prompt generation for a training image.}}

\vspace{-0.7cm}
\label{fig_TextPrompter}
\end{figure}

To incorporate domain-invariant textual information into the training process, we first design a Text Prompter (TP) module to dynamically supply matching and non-matching textual descriptions for images during training, based on the image label. As depicted in Fig.~\ref{fig_TextPrompter}, for each training image, the text prompter randomly selects a matching text prompt and a non-matching text prompt from the \textit{prompt library} for different types of images. For example, the matching text prompt to a print attack image can be \texttt{A man holding a printed photo.} while the non-matching text prompt could be \texttt{A photo of displaying a video clip on an iPad.} To ensure unbiased and diverse text prompts, we construct the \textit{prompt library} by employing GPT-4 \cite{openai2023gpt4} to automatically generate semantically similar phrases, eliminating subjective bias from manually designed text prompts. For instance, in the case of generating descriptive phrases for print attacks in Fig.~\ref{fig_TextPrompter}, we first manually craft an initial introductory text such as \texttt{A photo of a display} for the replay attack. Then, we query GPT-4 with text prompts like: \texttt{Write several short sentences with semantic similarity to \{A photo of a display.\}}. The generated phrases are then added to the \textit{prompt library} as textual descriptions of this image type (label). This approach provides diverse textual inputs to enhance the training effectiveness of the model (see Sec.~\ref{sec:ablation}).
\subsection{Methodology}
\label{sec:method}

\subsubsection{Hierarchical Attention Fusion Module}
\label{sec:method_part1}

\begin{figure*}[t]
\centering
\includegraphics[width=6in]{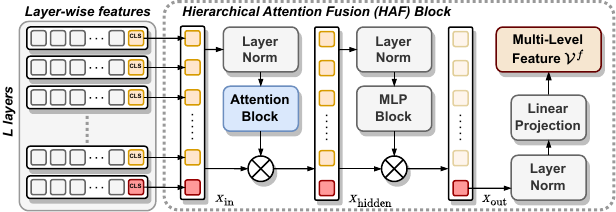}
\DeclareGraphicsExtensions.
\vspace{-0.1cm}
\caption{\textbf{Illustration of the designed Hierarchical Attention Fusion (HAF) module.} The proposed HAF is a lightweight plug-and-play module that can be easily integrated into mainstream ViT models.}
\vspace{-0.5cm}

\label{fig_LWA_block}
\end{figure*}

In face anti-spoofing, different attacks may lead to feature changes at different granularities. 
For instance, replay attacking involves more significant changes in the low-level texture features of the image, such as the moiré patterns~\cite{patel2015live}. On the other hand, attacks like printing or masking usually alter the high-level semantics of the image.
This necessitates the model to have the capability of aggregating features at different levels to cope with attacks at different scales. However, existing ViT models \cite{dosovitskiy2020image, liu2021swin, liu2022swin} typically use the \texttt{[CLS]} token of the last layer as the representation for recognition tasks, which makes it difficult to effectively integrate features at different levels, resulting in lower test accuracy. 

To solve this problem, we propose the Hierarchical Attention Fusion (HAF) module to achieve an adaptive merge of both the local texture features and the overall semantics of an input image. Experiments in Sec.~\ref{sec:ablation} empirically show that the designed HAF module significantly enhances ViTs in dealing with varied scales of attacks in FAS. Specifically, HAF fused information from \texttt{[CLS]} tokens of different layers in ViT through the self-attention block and the MLP block. 
Typically, for a training image $\boldsymbol{x}\in\mathbb{R}^{C\times H \times W}$, we first divide it into $N$ patches and feed it into a visual transformer with $L$-layers to obtain its layer-wise features $Z = \{\boldsymbol{z}^1, \boldsymbol{z}^2, \cdots, \boldsymbol{z}^L\}$, where $\boldsymbol{z} \in \mathbb{R}^{(N+1) \times D}$. A straightforward solution for feature aggregation is to perform self-attention on all features in $Z$. Nevertheless, it is computationally expensive and will bring many learnable parameters. Hence, we turn to utilizing the \texttt{[CLS]} token as an alternative for feature fusion. Since the information of $(N+1)$ token has been fused in the self-attention block of each layer, it can reduce redundancy and unnecessary operations compared to using all layer-wise features. Therefore, the input $X_{\text{in}} \in \mathbb{R}^{L \times D}$ of the HAF module is formulated as:

\vspace{-0.2cm}
\begin{equation}
\label{equation_CLStoken}    
X_{\text{in}} = (z^1_{\texttt{[CLS]}}, z^2_{\texttt{[CLS]}}, \cdots, z^L_{\texttt{[CLS]}})^T. 
\end{equation}

We first apply layer normalization (LN)~\cite{ba2016layer} to the layer-wise feature $X$ to alleviate internal covariate shifts caused by domain shifts. Then, the normalized features are projected with linear projections to align the dimensions of visual features with textual features, akin to the approach used in previous works~\cite{CLIP, BLIP, Lin_2023_CVPR} :
\begin{equation}
\label{equation_tQKV} 
Q = \mathrm{LN}(X_{\text{in}})W^Q \quad K = \mathrm{LN}(X_{\text{in}})W^K \quad V = \mathrm{LN}(X_{\text{in}})W^V
\end{equation}

where $ W^Q, W^K, W^V\in \mathbb{R}^{D \times D}$ are learned weight matrices for projecting to $Q$, $K$, and $V$ spaces respectively. Then the self-attention score is calculated as:
\vspace{-0.1cm}
\begin{equation}
\label{equation_AttentionBlock} 
\text{Attention}(Q,K,V) = \text{Softmax}\left(\frac{QK^T}{\sqrt{D}}\right)V,
\end{equation}
\vspace{-0.2cm}

where $D$ is the dimensionality of the key (and query) in the self-attention mechanism. We also follow the common practice of implementing multi-head  \cite{vaswani2017attention}. To enable the HAF module for capturing more complex features and patterns, we further add a multilayer perceptron (MLP) and residual connections~\cite{he2016deep} subsequently: 
\vspace{-0.2cm}
\begin{align}
\label{X_calc} 
\left\{
\begin{array}{l}
    X_{\text{hidden}} = \text{Attention}(Q,K,V)) + X_{\text{in}}\\
    X_{\text{out}} = \text{MLP}(\mathrm{LN}(X_{\text{hidden}})) + X_{\text{hidden}}
\end{array}
\right.
\end{align}
\vspace{-0.2cm}

Finally, we extract the normalized fused vision features from the last layer and map its dimension to the text feature with a learnable projection matrix $M\in \mathbb{R}^{D_{\text{out}} \times D}$ for cross-modal alignment (will be introduced in Sec. \ref{sec:method_part2}):

\vspace{-0.2cm}
\begin{equation}
\mathcal{V}^{f} = \mathrm{LN}(X_{\text{out}}^{(L)}) \cdot M.
\end{equation}
\vspace{-0.32cm}

The obtained visual feature $\mathcal{V}^{f}$ is aggregated with multi-level features from each layer of the ViT model, thereby capable of detecting various attacks at different granularities.

\subsubsection{Textual-Enhanced Visual Discriminator}
\label{sec:method_part2}

\begin{figure}[t]
\centering
\includegraphics[width=3in]{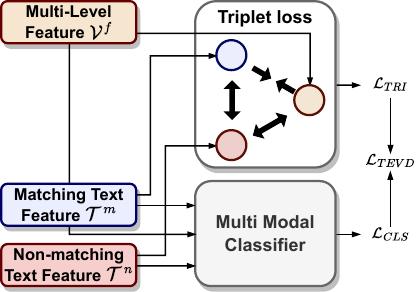}
\caption{\textbf{The proposed textual-enhanced visual discriminator (TEVD).} TEVD consists of a vision-language triplet loss and a multi-modal classifier.}

\vspace{-0.5cm}
\label{fig_Discriminator}
\end{figure}
%\vspace{-0.2cm}

To effectively integrate the text information into the training process, we further design a Textual-Enhanced Visual Discriminator (TEVD), which includes a vision-language triplet loss and a multi-modal classifier. We also make a detailed discussion and comparison between the proposed TEVD and the widely-used CLIP~\cite{CLIP} strategy in Tab.~\ref{tab:few_shot_perform}.

\textbf{Vision-Language Triplet loss.} 
We return to our core insight -- text can naturally capture the commonalities and essential characteristics across various attacks, thus bridging the gap between different image domains. To learn domain-invariant visual features, we leverage text features as supervision, guiding the model to filter out the inherent noise and domain-specific features (e.g., illumination, background) in visual representations. Concretely, a vision-language triplet loss is designed to achieve cross-modal alignment, where the text features can be interpreted as the prototypes of visual representations. It also results in more compact real-face features and distances the features of attack images from those of real ones, thereby improving the model's generalization ability to unknown domains. We also conduct studies when implementing the contrastive loss in the Appendix. 

To apply vision-language triplet loss, we first extract the visual multi-level feature $\mathcal{V}^{f}$ from input image $\boldsymbol{x}$ through our HAF module (introduced in Sec. \ref{sec:method_part1}), along with extracting the image's matching and non-matching text prompt feature $\mathcal{T}^m$ and $\mathcal{T}^n$ through text encoder. After L2 normalization, we form these features as a triplet $(\mathcal{V}^{f}_{\text{Norm}}, \mathcal{T}^m_{\text{Norm}}, \mathcal{T}^n_{\text{Norm}})$ for optimizing. The goal of the triplet loss $\mathcal{L}_{\text{TRI}}$ is to ensure that $\mathcal{V}^{f}_{\text{Norm}}$ is closer to $\mathcal{T}^m_{\text{Norm}}$ than $\mathcal{T}^n_{\text{Norm}}$:
%\vspace{-0.2cm}
\begin{equation}
\mathcal{L}_{\text{TRI}} = \max(0, D(\mathcal{V}^{f}_{\text{Norm}}, \mathcal{T}^m_{\text{Norm}}) - D(\mathcal{V}^{f}_{\text{Norm}}, \mathcal{T}^n_{\text{Norm}}) + \alpha),
\end{equation}

where $D(\mathcal{V}^{f}_{\text{Norm}}, \mathcal{T}^m_{\text{Norm}})$ denotes the Euclidean distance between $\mathcal{V}^{f}_{\text{Norm}}$ and $\mathcal{T}^m_{\text{Norm}}$, and $\alpha$ is a predefined margin.

\begin{algorithm}[t]
\caption{The Hierarchical Attention Fusion Module.}\label{alg:haf}
\begin{algorithmic}[1]
\State \textbf{Input:} Concatenate \texttt{[CLS]} tokens $X_{\text{in}}$.
\State \textbf{Output:} Extracted fused vision feature. $\mathcal{V}^{f}$
\Function {HAF\_Module}{$X_{\text{in}}$}
    \State \textcolor{CadetBlue}{\# Calculate features through attention and mlp}
    \State $X_{\text{hidden}}$ = Attention(LN($X_{\text{in}}$)) + $X_{\text{in}}$
    \State $X_{\text{out}}$ = MLP(LN($X_{\text{out}}$)) + $X_{\text{out}}$
    \State \textcolor{CadetBlue}{\# Project the vector dimensions to $D_{\text{out}}$}
    \State $\mathcal{V}^{f}$ = LN($X_{\text{out}}^{(L)}$) $\cdot$ M
    \State return $\mathcal{V}^{f}$
\EndFunction
\end{algorithmic}
\end{algorithm}

\begin{algorithm}[t]
\caption{The Textual-Enhanced Visual Discriminator.}\label{alg:vld}
\begin{algorithmic}[1]
\State \textbf{Input:} training images $x$, image binary label $y$, image types  $types$.
\State \textbf{Output:} TEVD module calculated loss $\mathcal{L}_{\text{TEVD}}$.
\Function {TEVD\_Module}{$x$, $y$, $types$}
    \State \textcolor{CadetBlue}{\# Generate matching and non-matching text prompts} 
    \State $P_m$, $P_n$ = TextPrompter($types$)
    \State \textcolor{CadetBlue}{\# Extract image and text features}
    \State $\mathcal{V}^{f}$ = VisionEncoder($x$)
    \State $\mathcal{T}^m$ = TextEncoder($P_m$)
    \State $\mathcal{T}^n$ = TextEncoder($P_n$)
    \State \textcolor{CadetBlue}{\# L2 normalize then concatenate }
    \State $\mathcal{V}^{f}_{\text{Norm}}$, $\mathcal{T}^m_{\text{Norm}}$, $\mathcal{T}^n_{\text{Norm}}$ = Normalize($\mathcal{V}^{f}$, $\mathcal{T}^m$, $\mathcal{T}^n$)
    \State $features$ = Cat(($\mathcal{V}^{f}_{\text{Norm}}$, $\mathcal{T}^m_{\text{Norm}}$, $\mathcal{T}^n_{\text{Norm}}$))
    \State $\text{labels}$ = Cat(($y$, $y$, Invert($y$)))
    \State \textcolor{CadetBlue}{\# Classification through multi-modal mlassifier $g$}
    \State $scores$ = g($features$)[:, 1] 
    \State \textcolor{CadetBlue}{\# Cross entropy loss and triplet loss}
    \State $\mathcal{L}_{\text{CLS}}$ = CrossEntropyLoss($scores$, $\text{labels}$)
    \State $\mathcal{L}_{\text{TRI}}$ = TripletsLoss($\mathcal{V}^{f}_{\text{Norm}}$, $\mathcal{T}^m_{\text{Norm}}$, $\mathcal{T}^n_{\text{Norm}}$)
    \State $\mathcal{L}_{\text{TEVD}}$ = $\mathcal{L}_{\text{CLS}}$ + $\lambda$ $\mathcal{L}_{\text{TRI}}$
    \State return $\mathcal{L}_{\text{TEVD}}$
\EndFunction
\end{algorithmic}
\end{algorithm}

\textbf{Multi-Modal Classifier.} To enhance the domain generalization ability in FAS, we hope to learn a classifier that can make reasonable judgments based on the consistent features across domains, and automatically discard the noise or the domain-specific details that may exist in the visual representations. To this end, we design a multi-modal classifier by classifying visual representations as well as corresponding matching and non-matching text prompt features simultaneously. The multi-modal classifier enriches the representation of the visual feature space and aligns visual features with the textual representation feature space. In other words, since the low-level textures and noise information are not involved in the pre-trained text feature, it serves as the regularization term to prevent overfitting of the classifier. Note that the language modality is removed during inference.

To better align the textual and visual features, we ensure that the prompt features corresponding to the input image labels are classified into the same category, while the non-matching prompt features are classified into the opposite category. Consider the binary label $y$ of the input image $\boldsymbol{x}$, we use $\neg y$ to represent the opposite label of $y$, and the multi-modal classifier is represented as $g$. The optimization objective is as follows: 
%\vspace{-0.2cm}
\begin{align}
\mathcal{L}_{\text{CLS}}=-(y\log(g(\mathcal{V}^{f}_{\text{Norm}}))+\neg y\log(1-g(\mathcal{V}^{f}_{\text{Norm}})))\notag\\
-(y\log(g(\mathcal{T}^m_{\text{Norm}}))+\neg y\log(1-g(\mathcal{T}^m_{\text{Norm}})))  \notag\\
-(y\log(1-g(\mathcal{T}^n_{\text{Norm}}))+\neg y\log(g(\mathcal{T}^n_{\text{Norm}})))
\end{align}
%\vspace{-0.2cm}

In summary, \texttt{TeG-DG} first applies the hierarchical attention fusion module to obtain enhanced multi-level visual representations, and further regularize the learned decision boundaries by the designed textual-enhanced visual discriminator. The overall loss function of \texttt{TeG-DG} during training is as follows:

%\vspace{-0.2cm}
\begin{equation}
\mathcal{L}_{\texttt{TeG-DG}} = \mathcal{L}_{\text{CLS}} + \lambda \mathcal{L}_{\text{TRI}},
\end{equation}
where $\lambda$ is a hyper-parameter. The PyTorch-style pseudo-code of HAF and TEVD is summarized in Alg.~\ref{alg:haf} and Alg.~\ref{alg:vld}. The algorithm of \texttt{TeG-DG} is presented in the Appendix.

\section{Experiment}
\label{sec:experiment}
\subsection{Experimental Settings}

\textbf{Datasets.} We evaluate the proposed method across four publicly available datasets~\cite{boulkenafet2017oulu, chingovska2012effectiveness, wen2015face, zhang2012face}, each manifesting significant domain variations (\emph{e.g.}, illumination, resolution, capture device, \emph{etc.}), especially for the spoof ones, which is highly suitable for DG-FAS experiments. These datasets are OULU-NPU~\cite{boulkenafet2017oulu} (denoted as O), Replay-Attack~\cite{chingovska2012effectiveness} (denoted as I), MSU-MFSD~\cite{wen2015face} (denoted as M), and CASIA-MFSD~\cite{zhang2012face} (denoted as C). To ensure a \textit{fair} comparison, we follow the same protocols employed by previous DG-based FAS methods~\cite{SSDG_FAS, liu2021adaptive, liu2021dual, shao2019multi, shao2020regularized, zhou2022adaptive, IADG_FAS} in all our experiments.

\textbf{Evaluation Metrics.} We use the Half Total Error Rate (HTER)~\cite{chingovska2014biometrics} and the Area Under Curve (AUC) for evaluation. HTER is a metric to evaluate the performance of a face anti-spoofing model, which is the average of the false acceptance rate (FAR) and the false rejection rate (FRR), measures the proportion of spoofing attacks and genuine faces that are incorrectly classified by the system. AUC aims to evaluate the performance of a binary classifier. It measures the area under the Receiver Operating Characteristic (ROC) curve, which plots the true positive rate against the false positive rate at different threshold values. 

\subsection{Comparisons to State-of-the-Art}

\setlength\tabcolsep{6.6pt}

\begin{table*}[t]
\centering
\caption{\textbf{Test HTER ($\downarrow$) and AUC ($\uparrow$) of FAS methods on OIMC datasets.} The * indicates using the CelebA-Spoof~\cite{CelebA-Spoof} as the supplementary source dataset (\textbf{bold} indicates best performance, \underline{underline} indicates second best performance.)}
\vspace{-0.3cm}
\begin{tabular}{l|cc|cc|cc|cc}
    \toprule
    \multicolumn{1}{l|}{\multirow{2}{*}{\textbf{Methods}}} & \multicolumn{2}{c|}{\small{\textbf{I\&C\&M to O}}} & \multicolumn{2}{c|}{\small{\textbf{O\&C\&M to I}}} & \multicolumn{2}{c|}{\small{\textbf{O\&C\&I to M}}} & \multicolumn{2}{c}{\small{\textbf{O\&M\&I to C}}} \\
     & HTER(\%) & AUC(\%) & HTER(\%) & AUC(\%) & HTER(\%) & AUC(\%) & HTER(\%) & AUC(\%) \\
    \midrule
    LBPTOP \cite{de2014face} & 53.15 & 44.09 & 49.45 & 49.54 & 36.90 & 70.80 & 42.60 & 61.05 \\
    MS\_LBP \cite{maatta2011face}& 50.29 & 49.31 & 50.30 & 51.64 & 29.76 & 78.50 & 54.28 & 44.98 \\
    MMD-AAE \cite{li2018domain}& 40.98 & 63.08 & 31.58 & 75.18 & 27.08 & 83.19 & 44.59 & 58.29 \\
    MADDG \cite{shao2019multi}& 27.98 & 80.02 & 22.19 & 84.99 & 17.69 & 88.06 & 24.50 & 84.51 \\
    RFM \cite{shao2020regularized}& 16.45 & 91.16 & 17.30 & 90.48 & 13.89 & 93.98 & 20.27 & 88.16 \\
    SSDG-M \cite{SSDG_FAS} & 25.17 & 81.83 & 18.21 & 94.61 & 16.67 & 90.47 & 23.11 & 85.45 \\
    D\textsuperscript{2}AM \cite{chen2021generalizable}& 15.27 & 90.87 & 15.43 & 91.22 & 12.70 & 95.66 & 20.98 & 85.58 \\
    DRDG \cite{liu2021dual}& 15.63 & 91.75 & 15.56 & 91.79 & 12.43 & 95.81 & 19.05 & 88.79 \\
    ANRL \cite{liu2021adaptive}& 15.67 & 91.90 & 16.03 & 91.04 & 10.83 & 96.75 & 17.85 & 89.26 \\
    SSAN-M \cite{SSAN_FAS} & 19.51 & 88.17 & 14.00 & 94.58 & 10.42 & 94.76 & 16.47 & 90.81 \\
    AMEL \cite{zhou2022adaptive}& 11.31 & 93.96 & 18.60 & 88.79 & 10.23 & 96.62 & 11.88 & 94.39 \\
    DBDG \cite{du2022energy}& 15.66 & 92.02 & 18.69 & 92.28 & 9.56 & 97.17 & 18.34 & 90.01 \\
    SA-FAS \cite{sun2023rethinking} & 10.00 & 96.23 & 6.58 & 97.54 & 5.95 & 96.55 & 8.78 & 95.37 \\
    IADG \cite{IADG_FAS}& \underline{8.86} & \underline{97.14} & 10.62 & 94.50 & 5.41 & 98.19 & 8.70 & 96.44 \\
    DiVT-M \cite{liao2023domain}& 13.06 & 94.04 & \underline{3.71} & \underline{99.29} & \underline{2.86} & \underline{99.14} & \underline{8.67} & \underline{96.92} \\
    \rowcolor{graycolor} \texttt{TeG-DG} (ours) & \textbf{5.68} & \textbf{97.92} & \textbf{3.21} & \textbf{99.63} & \textbf{1.88} & \textbf{99.72} & \textbf{3.17} & \textbf{99.79}  \\
    \midrule
    FLIP-MCL* \cite{Srivatsan_2023_ICCV} & \textbf{2.31} & \underline{99.63} & \underline{4.25} & \underline{99.07} & \underline{4.95} & \underline{98.11} & \underline{0.54} & \underline{99.98} \\
    \rowcolor{graycolor} \texttt{TeG-DG}* (ours) & \underline{2.53} & \textbf{99.76} & \textbf{2.38} & \textbf{99.69} & \textbf{1.06} & \textbf{99.99} & \textbf{0.40} & \textbf{99.99}\\
    \bottomrule
\end{tabular}
\label{tab:LOO_performance}
\vspace{-0.5cm}
\end{table*}

\label{sec: leave_one_out}
\textbf{Leave-One-Out.} In alignment with previous researches~\cite{shao2019multi, SSDG_FAS, liu2021dual, liu2021adaptive, zhou2022adaptive, du2022energy, SSAN_FAS, IADG_FAS}, we adopted the same Leave-One-Out (LOO) protocol for our experiments. Detailed comparative results are thoroughly documented in Tab.~\ref{tab:LOO_performance}. Under the LOO protocol, three datasets are chosen to serve as source domains, while the remaining dataset is designated as the unseen target domain, not accessible during training. We compare different FAS methods in Tab.~\ref{tab:LOO_performance} including traditional FAS approaches~\cite{de2014face, maatta2011face} that exhibit subpar performance in domain generalization, as well as domain generalization FAS methods~\cite{li2018domain, shao2019multi, shao2020regularized, SSDG_FAS, chen2021generalizable, liu2021adaptive, SSAN_FAS, zhou2022adaptive, du2022energy, IADG_FAS, liao2023domain} that do not incorporate textual supervision. Our methodology, integrating textual supervision, outshines nearly all existing domain generalization frameworks under the LOO protocol. By aligning features with abstract textual space, our model, despite its simpler architecture, excels over more intricate systems that depend on feature whitening transformations and domain alignment and demonstrates significant improvements over these traditional methods.

\begin{table}[t]
\centering
\caption{\textbf{Results on extremely limited source domains.}}
\vspace{-0.3cm}
\setlength{\tabcolsep}{1.7pt}
\begin{tabular}{l|cc|cc}
    \toprule
    \multicolumn{1}{l|}{\multirow{2}{*}{\textbf{Methods}}} & \multicolumn{2}{c|}{\small{\textbf{M\&I to C}}} & \multicolumn{2}{c}{\small{\textbf{M\&I to O}}} \\
     & HTER(\%) & AUC(\%) & HTER(\%) & AUC(\%) \\
    \midrule
    LBPTOP \cite{de2014face} & 45.27 & 54.88 & 47.26 & 50.21 \\
    MS\_LBP \cite{maatta2011face}& 51.16 & 52.09 & 43.63 & 58.07 \\
    MADDG \cite{shao2019multi}& 41.02 & 64.33 & 39.35 & 65.10 \\
    SSDG-M \cite{SSDG_FAS} & 31.89 & 71.29 & 36.01 & 66.88 \\
    D\textsuperscript{2}AM \cite{chen2021generalizable}& 32.65 & 72.04 & 27.70 & 75.36 \\
    DRDG \cite{liu2021dual}& 31.28 & 71.50 & 33.35 & 69.14 \\
    ANRL \cite{liu2021adaptive}& 31.06 & 72.12 & 30.73 & 74.10 \\
    SSAN-M \cite{SSAN_FAS} & 30.00 & 76.20 & 29.44 & 76.62 \\
    EBDG \cite{du2022energy} & 27.97 & 75.84 & 25.94 & 78.28 \\
    AMEL \cite{zhou2022adaptive} & 24.52 & 82.12 & 19.68 & 87.01 \\
    IADG \cite{IADG_FAS}& 24.07 & 85.13 & \underline{18.47} & \underline{90.49}\\
    DiVT-M \cite{liao2023domain} & \underline{20.11} & \underline{86.71} & 23.61 & 85.73\\
    \midrule
    \rowcolor{graycolor}{\texttt{TeG-DG}} (ours) & \textbf{6.19} & \textbf{98.64} & \textbf{6.89} & \textbf{97.49} \\
    \bottomrule
\end{tabular}

\vspace{-0.6cm}
\label{tab:performance2}
\end{table}

\textbf{Limited source domains.} We also evaluate our method in scenarios with extremely limited source domains. Following previous works~\cite{shao2019multi, SSDG_FAS, liu2021adaptive, IADG_FAS}, MSU-MFSD (M) and ReplayAttack (I) are selected as the source domains for training, while the remaining CASIA-MFSD (C) and OULU-NPU (O) datasets will be used as the target domains for testing respectively. The result is shown in Tab.~\ref{tab:performance2}. Our \texttt{TeG-DG} method significantly outperforms state-of-the-art approaches, showing $\sim$\textbf{14\%} and $\sim$\textbf{12\%} improvements on HTER and AUC respectively, 
Notably, it outperforms most traditional domain generalization (DG) models, even those trained on a wider range of source datasets when testing on CASIA-MFSD. And even outperform all the methods we listed in Tab.~\ref{tab:LOO_performance} when testing on OULU-NPU. Our approach does not require domain labels and leverages the inherent abstract information provided by text, yielding better performance with fewer training domains. Given the practical challenges of acquiring sufficient source domain FAS data in real-world settings, the superior performance of our method with fewer source domains indicates the substantial value of text supervision for generalization in FAS tasks.

\subsection{Few-Shot Performance}
\label{sec:few_shot}

Gathering diverse attack samples for Facial Anti-Spoofing (FAS) tasks is strenuous and complex. Thus, it is extremely advantageous if we have a model with robust few-shot learning capabilities. In contrast to previous few-shot methods which rely on training models on known datasets and then performing few-shot learning on unseen domains~\cite{qin2020learning, perez2020learning}, we have adopted a more stringent Leave-One-Out (LOO) few-shot protocol, similar to the one detailed in Section~\ref{sec: leave_one_out}. However, we differ in that we utilize only a limited number of images for training and employ a zero-shot approach for the unseen domain.
To demonstrate the effectiveness of our approach in zero/few-shot learning, we conducted a comparison with ``ViT-L/14" CLIP~\cite{CLIP}. The results are shown in Tab.~\ref{tab:few_shot_perform}, where `$n$ sample' denotes $n$ spoofing and $n$ real samples from each source domain, and `zero-shot' indicates no image samples given. For the zero-shot scenario, our method only activates the multi-modal classifier, and CLIP classifies only via text prompts. For few-shot learning, we use a linear probe to finetune the classification head of CLIP. As training samples increase, both our method and CLIP show an expected overall improvement in classification ability. However, our approach consistently outperforms CLIP. It is also noteworthy that our model, using only a few samples from the source domain, achieved performance surpassing most DG FAS methods in Tab.~\ref{tab:LOO_performance} which completely use the whole source domain. Our superior results demonstrate the value of utilizing textual information for generalization.

\setlength\tabcolsep{6pt}
\begin{table*}[t]
\centering
\caption{\textbf{Zero-shot and few-shot performance evaluated on Leave-One-Out (LOO) protocol.}}
\vspace{-0.3cm}
\begin{tabular}{l|cc|cc|cc|cc}
    \toprule
    \multicolumn{1}{l|}{\multirow{2}{*}{\textbf{Methods}}} & \multicolumn{2}{c|}{\small{\textbf{I\&C\&M to O}}} & \multicolumn{2}{c|}{\small{\textbf{O\&C\&M to I}}} & \multicolumn{2}{c|}{\small{\textbf{O\&C\&I to M}}} & \multicolumn{2}{c}{\small{\textbf{O\&M\&I to C}}} \\
     & HTER(\%) & AUC(\%) & HTER(\%) & AUC(\%) & HTER(\%) & AUC($\uparrow$) & HTER(\%) & AUC(\%) \\
    \midrule
    CLIP (zero-shot) & 60.35 & 37.23 & 58.73 & 39.09 & 51.84 & 42.12 & 51.02 & 50.12\\
    \rowcolor{graycolor}{\texttt{TeG-DG}} (zero-shot) & \textbf{41.05} & \textbf{61.67} & \textbf{34.44} & \textbf{68.57} & \textbf{22.69} & \textbf{78.17} & \textbf{38.64} & \textbf{61.93}\\
    \midrule
    CLIP (1 sample) & 16.48 & 90.64 & 18.21 & 88.41 & \textbf{22.24} & \textbf{88.73} & 32.45 & 76.02\\
    \rowcolor{graycolor}{\texttt{TeG-DG}} (1 sample) & \textbf{6.34} & \textbf{98.45} & \textbf{13.16} & \textbf{95.22} & 27.88 & 85.56 & \textbf{7.70} & \textbf{98.26}\\
    \midrule
    CLIP (5 sample) & 11.74 &  94.43 &  16.99 &  90.11 &  10.22 &  96.86 &  19.85 & 89.75 \\
    \rowcolor{graycolor}{\texttt{TeG-DG}} (5 sample) & \textbf{11.32}  &  \textbf{95.38} &  \textbf{9.69} & \textbf{97.01}  &  \textbf{5.56} &  \textbf{99.01} & \textbf{4.16}  & \textbf{99.59} \\
    \bottomrule
\vspace{-0.3cm}
\end{tabular}

\label{tab:few_shot_perform}
\end{table*}

\subsection{Analysis and Ablation Study}
\label{sec:ablation}

\textbf{The effectiveness of proposed components.} We conducted an ablation study to evaluate the performance contribution of each component in our framework, \emph{i.e.}, the HAF module (denoted as HAF), the triplet loss in the TEVD module (denoted as triplet), and the textual guidance in TEVD module (denoted as text). The results are shown in Tab.~\ref{tab:ablation_experiment}. It's observed that a total loss of textual information will harm the performance more than the absence of the triplet loss in TEVD. The HAF module effectively enhances the model's performance on datasets with lower image quality, demonstrating its improved capability in feature extraction across images of varying quality. Each component of \texttt{TeG-DG} contributes to improving the performance across various architectures. Finally, the combination of all the proposed components yields the best results.

\begin{table*}[t]
\centering
\caption{\textbf{Evaluations of different components of the proposed \texttt{TeG-DG} framework.}}
\vspace{-0.3cm}
\begin{tabular}{l|cc|cc|cc|cc}
    \toprule
    \multicolumn{1}{l|}{\multirow{2}{*}{\textbf{Methods}}} & \multicolumn{2}{c|}{\small{\textbf{I\&C\&M to O}}} & \multicolumn{2}{c|}{\small{\textbf{O\&C\&M to I}}} & \multicolumn{2}{c|}{\small{\textbf{O\&C\&I to M}}} & \multicolumn{2}{c}{\small{\textbf{O\&M\&I to C}}} \\
     & HTER(\%) & AUC(\%) & HTER(\%) & AUC(\%) & HTER(\%) & AUC($\uparrow$) & HTER(\%) & AUC(\%) \\
    \midrule
    {\texttt{TeG-DG}} w/o HAF &  9.86 &  95.87 & 13.16 & 94.35 & 3.23 & 99.51 & 14.34 & 92.26 \\
    {\texttt{TeG-DG}} w/o triplet & 11.65 &  94.75 &  9.44 & 97.40 & 5.11 &  99.02 &  8.68 & 97.40 \\
    {\texttt{TeG-DG}} w/o text & 10.71 & 94.90 & 16.13 & 89.99 & 6.99 & 97.50 & 7.09 & 97.92 \\
    \rowcolor{graycolor}{{\texttt{TeG-DG}} (ours)} & \textbf{5.68} & \textbf{97.92} & \textbf{7.65} & \textbf{98.67} & \textbf{1.88} & \textbf{99.72} & \textbf{3.17} & \textbf{99.79}  \\
    \bottomrule
\end{tabular}
\vspace{-0.3cm}
\label{tab:ablation_experiment}
\end{table*}

\begin{figure}[t]
\centering
\includegraphics[width=3.25in]{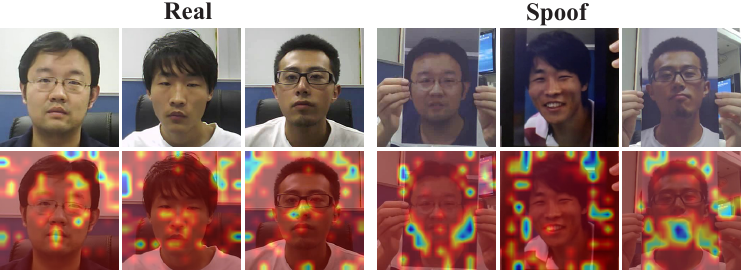}
%\vspace{-0.4cm}
\caption{\textbf{The Grad-CAM \cite{selvaraju2017grad} visualizations of our \texttt{TeG-DG} method under protocol O\&M\&I to C.} }
%\vspace{-0.2cm}
\label{fig_gradcam}
\end{figure}

\begin{figure}[t]
\centering
\includegraphics[width=3.25in]{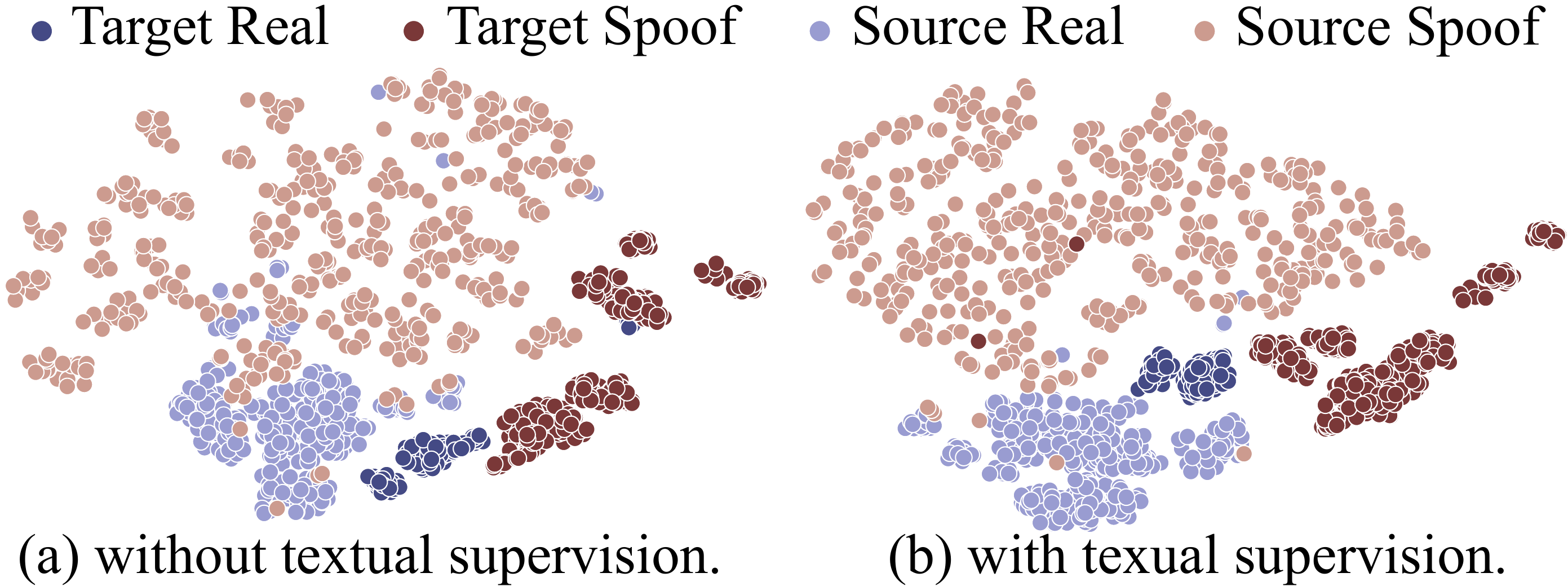}
\caption{\textbf{The t-SNE feature visualization on O\&M\&I to C.} We plot the visual feature distribution w/ and w/o \texttt{TeG-DG}.}

%\vspace{-0.6cm}
\label{fig_tsne}
\end{figure}

\textbf{Grad-CAM Visualization.} To provide a better understanding of \texttt{TeG-DG}, we further identify the regions that influence the classification results. We used Grad-CAM~\cite{selvaraju2017grad} for activation map visualization on input images. The left side of Fig.~\ref{fig_gradcam} shows the real images, while the right side shows the spoof ones. The first line and second lines refer to the input images and the corresponding activation maps, respectively. The proposed method primarily emphasizes the internal regions of real faces as cues for classification, while for attack images, the focus is more scattered to attack indicators such as photo edges and screen-caused features. This demonstrates the strong capability of \texttt{TeG-DG} to distinguish real and counterfeit features in unseen domains.

\textbf{t-SNE Visualization.} Figure~\ref{fig_tsne} displays the t-SNE feature visualization for protocol O\&M\&I to C \cite{van2008visualizing} with and without \texttt{TeG-DG}, showcasing the impact of textual supervision information. These textual supervisions facilitate a more pronounced separation in the feature space between real and spoof faces, simultaneously achieving denser and more generalized clustering of real face features. In the absence of textual supervision, the distinction between real and spoof features within the source and target domain becomes blurred. This contrast is especially noticeable when comparing the separation extent in target domain features with and without textual supervision. The t-SNE visualization of the \texttt{TeG-DG} feature distribution demonstrates the enhanced generalization capability of our framework, highlighting the clear advantages of integrating textual supervision.

\section{Conclusion and Limitations}
\label{sec:conclusion}

In this paper, we introduced the Textually Guided Domain Generalization (\texttt{TeG-DG}) framework which utilizes the natural cross-domain universality of text and leverages it to bridge the gap between various visual domains, thereby enhancing domain generalization performance. We develop the Hierarchical Attention Fusion (HAF) module to merge both the local and the high-level semantics in the visual modality and a Textual-Enhanced Visual Discriminator (TEVD) incorporates textual supervision in the language modality. 
Comprehensive experiments on benchmark datasets demonstrate the superiority of our approach. There are nevertheless some limitations. First, our method requires labeling spoofing attack types, which may sometimes necessitate special annotations. Second, the creation of the \textit{prompt library} would impact the performance. We hope our work can promote the research on domain generalization FAS tasks.

% {\small
% \bibliographystyle{ieeenat_fullname}
% \bibliography{sections/11_references}
% }

\ifarxiv \clearpage \appendix \section{Data Preparation}

\subsection{Dataset Construction}
In alignment with standard practices in Face Anti-Spoofing (FAS) research, we adopted the same Leave-One-Out (LOO) protocol for our experiments. Under the LOO protocol, three datasets are chosen to serve as source domains, while the remaining dataset is designated as the unseen target domain, not accessible during training. We also evaluate our method in scenarios with extremely limited source domains. The dataset for LOO test and extremely limited source domains test is listed in Tab.~\ref{tab:dataset_intro}.
During the training phase, we maintain a balanced approach by sampling an equal number of real and spoof samples. This balance is crucial for training our system to accurately distinguish between real and spoof images under varied conditions. The datasets employed in our study are:

\textbf{OULU-NPU:} This dataset mimics real-world variations in mobile face presentation attacks, encompassing diverse environmental conditions, lighting variations, and a range of mobile devices.

\textbf{Replay-Attack:} This dataset offers an extensive collection of video replay attacks, making it invaluable for testing the effectiveness of low level features in identifying such spoofing attempts.

\textbf{MSU-MFSD:} This dataset focused on face spoof detection using image distortion analysis, it includes a variety of spoof attacks, highlighting the importance of image quality and distortion in detecting fraud.

\textbf{CASIA-MFSD:} This dataset is renowned for its diverse spoofing attacks, it includes methods like warped photo attacks, cut photo attacks, and video replays, providing a comprehensive assessment platform for anti-spoofing systems.

\begin{table}[h]
\centering
\caption{\textbf{Four datasets for Leave-One-Out test.}}
\vspace{-0.2cm}
\label{tab:dataset_intro}
\begin{tabularx}{\columnwidth}{ccc}
\hline
\textbf{Dataset} & \textbf{Live/Spoof} & \textbf{Attack Types} \\ \hline
CASIA-MFSD~\cite{zhang2012face} & 150/450 & Print, Replay\\ \hline
REPLAY-ATTACK~\cite{chingovska2012effectiveness}& 200/1000 & Print, Replay\\ \hline
MSU-MFSD~\cite{wen2015face} & 70/210 & Print, Replay\\ \hline
OULU-NPU~\cite{boulkenafet2017oulu} & 720/2880 & Print, Replay\\ \hline
\end{tabularx}
\end{table}

\subsection{Prompts Preparation}
\begin{algorithm}[t]
\caption{The train process of \texttt{TeG-DG}.}\label{alg:overall}

\textbf{Input:} Train set $\mathcal{D}$, hyper-parameter $\lambda$, train epoch $T$, predefined prompt library $PL$.\\
\textbf{Output}: Trained model parameter $\theta_T$.\\
\textbf{Initialize}: Load ``ViT-L/14" CLIP \cite{CLIP} pre-trained parameters for the backbone in as the backbone for both the vision and language modalities of our model. The classifier in the TEVD module and parameters in the HAF module are randomly initialized.
\begin{algorithmic}[1]
\State Build up LOO dataset from $\mathcal{D}$
\State Load pre-trained parameters to model
\State Freeze backbone model parameters except for \texttt{ln\_pre} and \texttt{ln\_post} in ViT
\For {$\text{epoch} = 0,\cdots,T-1$}
    \State Get training images $x$, binary label $y$ and image types $types$ from $\mathcal{D}$, and do pre-processing
    \State Train model parameter $\theta_T$ though $\mathcal{L}_{\text{TEVD}}$ (Sec.~\ref{sec:method_part2})
\EndFor
\State Save model parameter $\theta_T$
\end{algorithmic}
\end{algorithm}

\begin{algorithm}[t]
\caption{The test process of \texttt{TeG-DG}.}\label{alg:infer}
\textbf{Input:} Unseen domain test set $\mathcal{D}^t$, trained model parameter $\theta_T$\\
\textbf{Output}: HTER and AUC (Sec.~\ref{sec:experiment}).
\begin{algorithmic}[1] % The number tells where the line numbering should start
\State $ScoreList$ = []
\State $LabelList$ = []
\For {$x_i, y_i \in \mathcal{D}^t$}
    \State Obtain the classification score $s_i$ of $x_i$ via the visual part of the model (Sec.~\ref{sec:method_part1}). 
    \State Add $s_i$ to $ScoreList$
    \State Add $y_i$ to $LabelList$
\EndFor
\State Evaluate model with HTER and AUC through $ScoreList$ and $LabelList$
\end{algorithmic}
\end{algorithm}

We construct the \textit{prompt library} by employing GPT-4 \cite{openai2023gpt4} to generate semantically similar phrases automatically. We first manually craft an initial introductory text such as \texttt{A photo of a display} for the replay attack. Then, we query GPT-4 with text prompts like: \texttt{Write several short sentences with semantic similarity to \{A photo of a display.\}}. We query GPT-4 with the following text for batch generation: \texttt{Write \{64\} numbered sentences with semantic similarity to \{A photo of a display.\}}. In this way, we generate 64 text prompts for each type of image included in adopted datasets. The generated text prompts are then added to the \textit{prompt library} as textual descriptions of this image type. The text prompts used in the \textit{prompt library} are listed in Page~\pageref{text_prompts}. During training, these will be dynamically selected as matching or non-matching text prompts to pair with training images.

\section{Pseudo-code of \texttt{TeG-DG} Framework}

Our \texttt{TeG-DG} framework is designed to simultaneously engage both visual and language modalities during the training phase to max out domain generalization ability, offering a multi-faceted approach to learning. Key to this process is the Hierarchical Attention Fusion (HAF) module, which adeptly merges local texture features with high-level semantics extracted from input images. The HAD module is detailed in Sec.~\ref{sec:method_part1}, with its pseudo-code presented in Alg.~\ref{alg:haf}.

In tandem with HAF, our framework features the Textual-Enhanced Visual Discriminator (TEVD), a crucial component in tackling the challenges of domain generalization through textual supervision. The TEVD's role and functionalities are extensively discussed in Sec.~\ref{sec:method_part2}, and its pseudo-code is presented in Alg.~\ref{alg:vld}. The training phase of the visual modality in our framework not only utilizes real/spoof image labels but also incorporates textual information to enrich the learning process.

This multi-modal training approach is encapsulated in Alg.~\ref{alg:overall}, which details the pseudo-code for the entire training procedure. And during the inference stage, our framework pivots to focus exclusively on the visual modality. The language modality, pivotal during training, is set aside in this phase. For inference purposes, we utilize a multi-modal classifier, which has been trained with text regularization in the TEVD module. This inference process is detailed in Alg.\ref{alg:infer}, outlining the complete pseudo-code of inference process.

\section{More Experiment and Analysis}
\subsection{Comparison to Baseline Method}

\begin{figure}[t]
\centering
\includegraphics[width=3.25in]{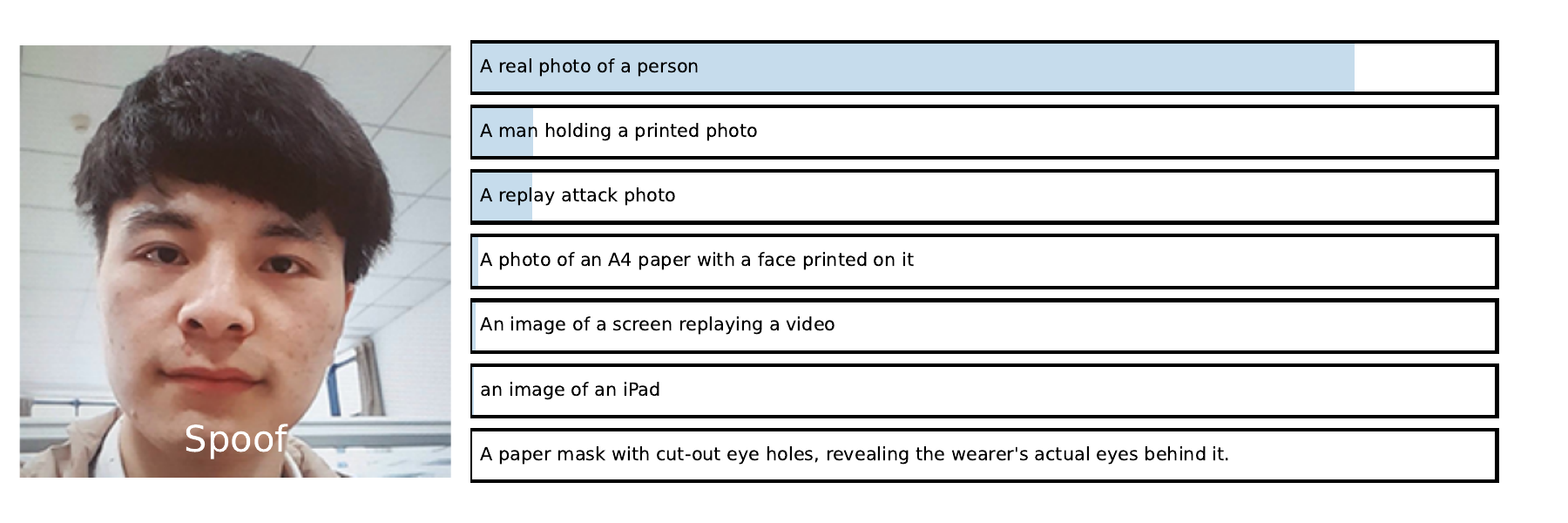}

\vspace{-0.1cm}
\includegraphics[width=3.25in]{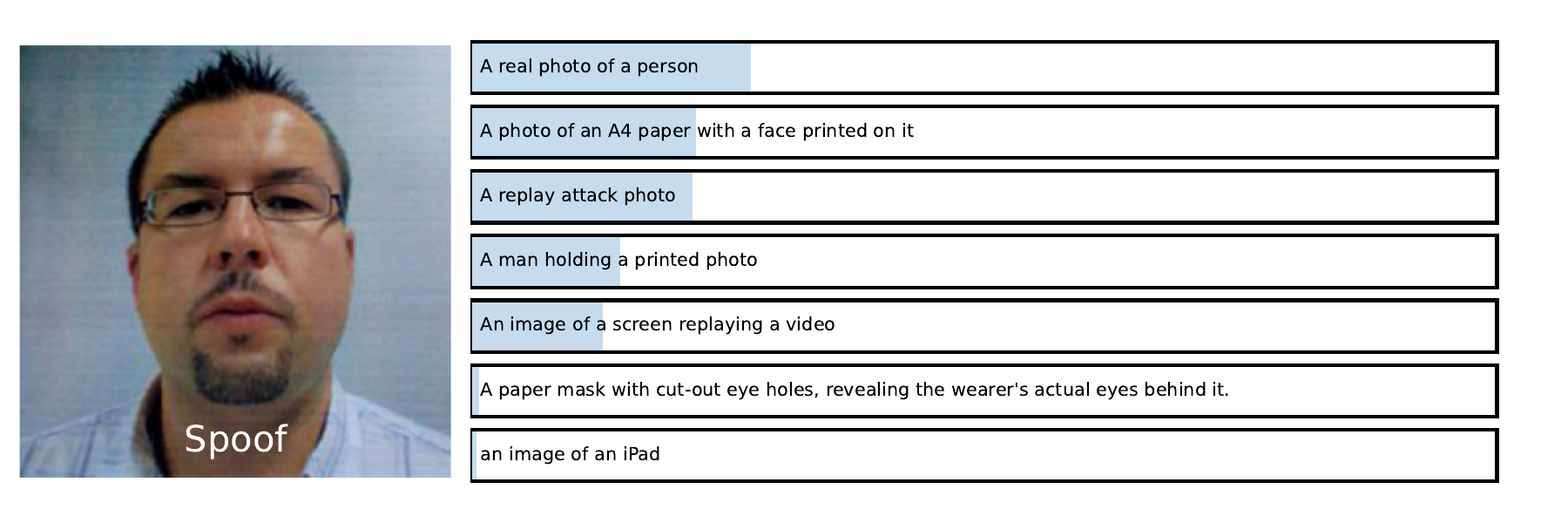}

\vspace{-0.1cm}
\includegraphics[width=3.25in]{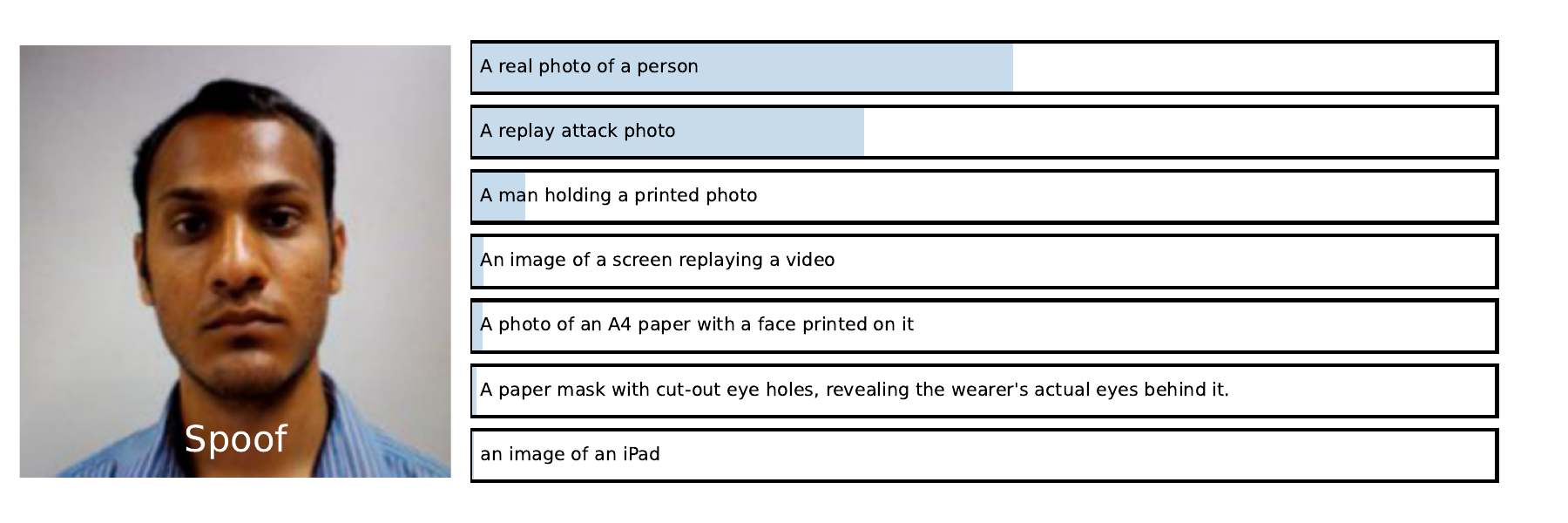}

\vspace{-0.1cm}
\includegraphics[width=3.25in]{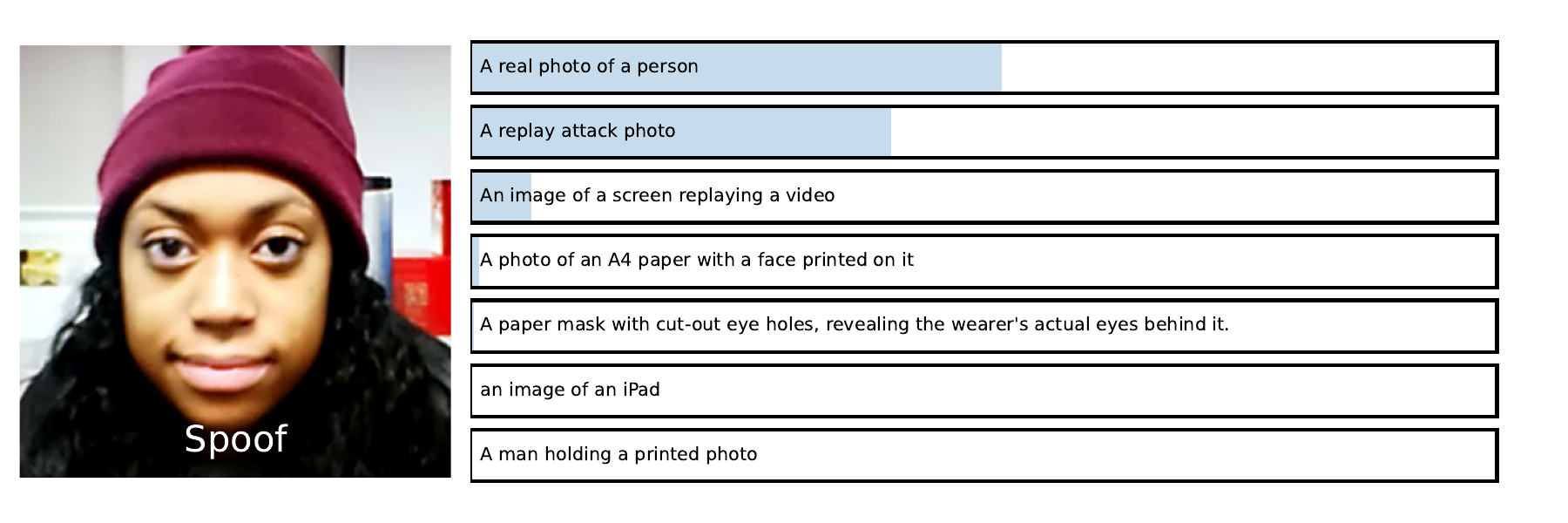}

\vspace{-0.1cm}
\includegraphics[width=3.25in]{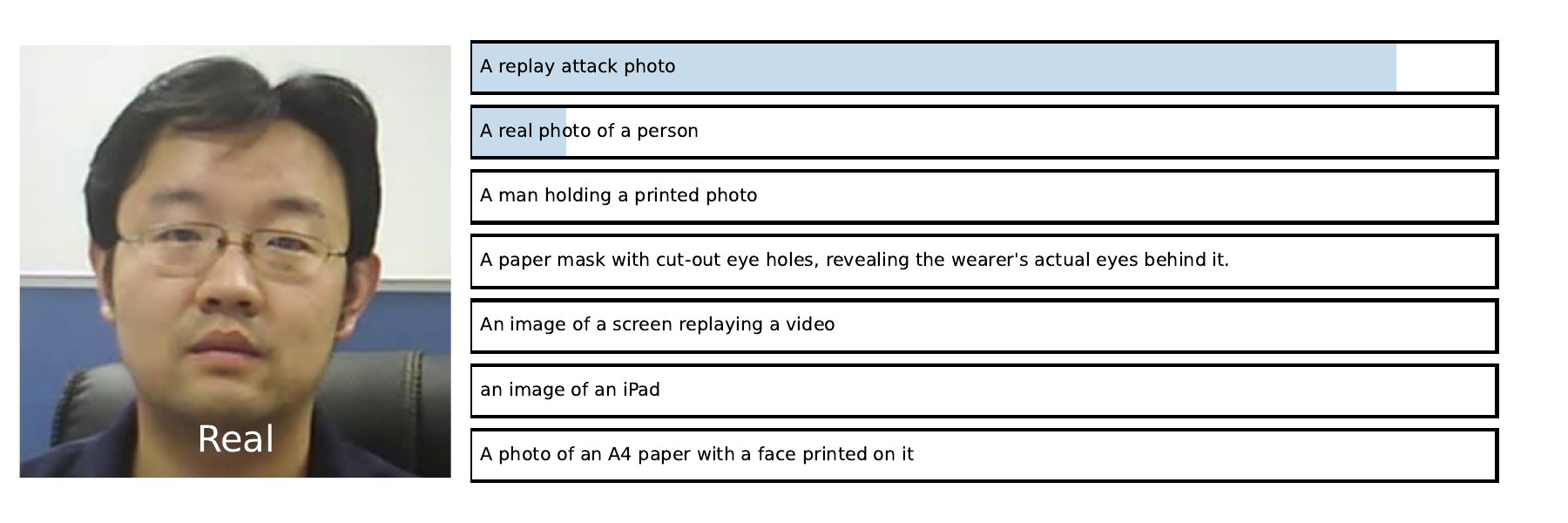}

\vspace{-0.1cm}
\includegraphics[width=3.25in]{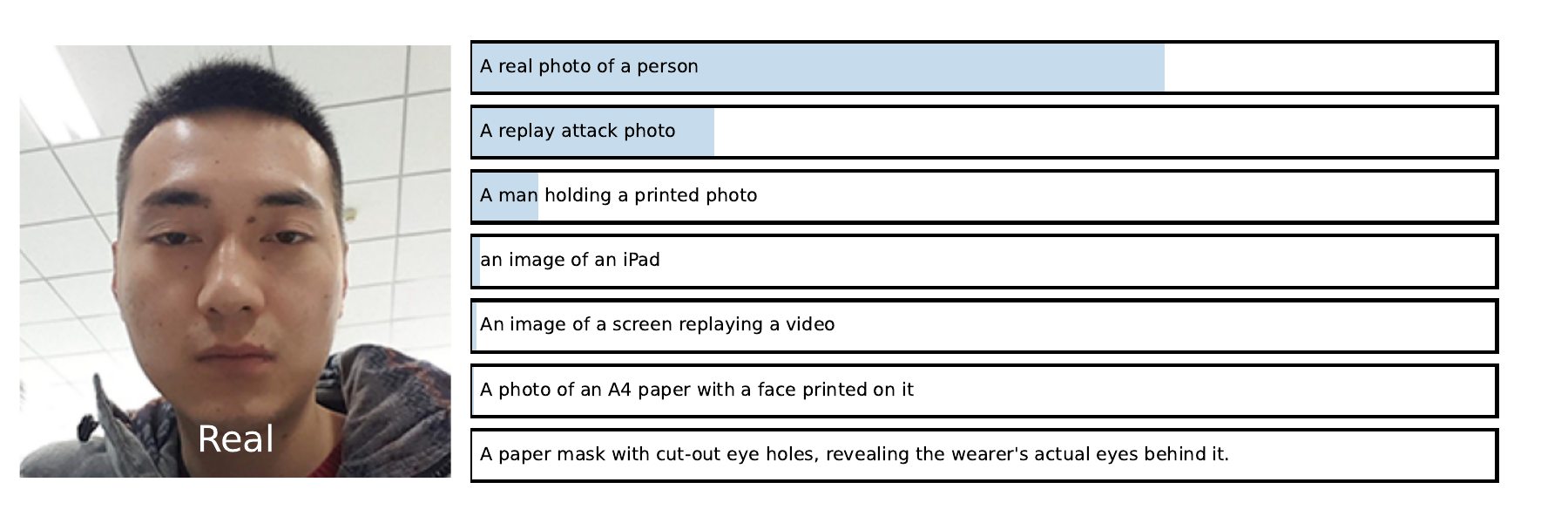}

\vspace{-0.1cm}
\caption{\textbf{CLIP model lacks the knowledge of FAS tasks.}}
%\vspace{-0.6cm}
\label{fig:clip_scores}
\end{figure}

Our \texttt{TeG-DG} framework utilizes a pre-trained Visual Language Model (VLM), but the VLM model that was not specifically designed for FAS tasks performs poorly. For example, the CLIP model exhibits a lack of specific knowledge pertinent to Face Anti-Spoofing (FAS) tasks. In the context of the four given datasets, it tends to mistakenly classify images with sharp visuals as authentic and those with blurred details as attacks, thereby overlooking other critical FAS features such as moiré patterns and loss of facial details as shown in Fig.~\ref{fig:clip_scores}. This oversight leads to misjudgments. Consequently, CLIP in its standalone form underperforms in FAS tasks.

To validate the effectiveness of our approach compared to the direct application of Vision-Language Models (VLMs), we used CLIP\cite{CLIP} as a baseline for comparison. We added BLIP\cite{BLIP} for more comprehensive measurement. Our experiments demonstrate that our method holds advantages over several alternative strategies, including using a VLM model directly, fine-tuning a VLM with a classification head, or employing contrastive loss fine-tuning with a VLM. These results are listed in Tab.~\ref{tab:baseline_compare} and underscore the superiority of our methodology in leveraging vision-language synergies.

\setlength\tabcolsep{7pt}

\begin{table*}[t]
\centering
\caption{\textbf{Comparison to Baseline Method.} `(Linear)' means only using a linear classification head.}
%\vspace{-0.3cm}
\begin{tabular}{l|cc|cc|cc|cc}
    \toprule
    \multicolumn{1}{l|}{\multirow{2}{*}{\textbf{Methods}}} & \multicolumn{2}{c|}{\small{\textbf{I\&C\&M to O}}} & \multicolumn{2}{c|}{\small{\textbf{O\&C\&M to I}}} & \multicolumn{2}{c|}{\small{\textbf{O\&C\&I to M}}} & \multicolumn{2}{c}{\small{\textbf{O\&M\&I to C}}} \\
     & HTER(\%) & AUC(\%) & HTER(\%) & AUC(\%) & HTER(\%) & AUC($\uparrow$) & HTER(\%) & AUC(\%) \\
    \midrule
    CLIP (zero-shot) & 60.35 & 37.23 & 58.73 & 39.09 & 51.84 & 42.12 & 51.02 & 50.12\\
    CLIP (Linear) & 11.03 & 94.80 & 12.50 & 93.97 & 2.78 & 99.69 & 8.68 & 97.18\\
    BLIP (Linear) & 23.21 & 80.82 & 16.93 & 87.23 & 12.87 & 94.26 & 27.45 & 78.71\\
    contrastive Loss & 6.54 &  98.07 &  8.29 &  98.66 &  3.84 &  99.65 &  7.24 & 98.20 \\
    \rowcolor{graycolor} \texttt{TeG-DG} (ours) & \textbf{5.68} & \textbf{97.92} & \textbf{3.21} & \textbf{99.63} & \textbf{1.88} & \textbf{99.72} & \textbf{3.17} & \textbf{99.79}  \\
    \bottomrule
\end{tabular}
\label{tab:baseline_compare}
\end{table*}
\setlength\tabcolsep{6pt}

\subsection{Impact of Different Numbers of Prompts}
\label{appen: prompt_numbers}

Our framework employs a prompt library to store currently used, predefined text prompts, and the number of these prompts can significantly impact the achievable results. To explore the effects of using different numbers of prompts, we extracted $n$ text prompts for each image category to perform Leave-One-Out (LOO) tests, investigating the variations in Half Total Error Rate (HTER) and Area Under the Curve (AUC) to determine the influence of varying prompt quantities. The experimental results are shown in Fig.~\ref{fig:prompt_ICM_O}, Fig.~\ref{fig:prompt_OCM_I}, Fig.~\ref{fig:prompt_OCI_M} and Fig.~\ref{fig:prompt_OMI_C}. It reveals that in most cases, employing a larger number of prompts tends to lower the HTER and increase the AUC. This improvement is attributed to the richer textual feature space, which overall enhances the model's generalization performance. The appropriate quantity of text prompts is beneficial within our framework, and selecting an optimal number of text prompts is essential for maximizing the performance within our framework. Based on these findings, we chose to use 64 text prompts in our previous experiments. This decision was made to leverage the benefits of a richer textual feature space while maintaining operability and effectiveness in our model's performance.

\section{Visualization}

We provide more Grad-Cam~\cite{selvaraju2017grad} visualization \texttt{TeG-DG} in Fig.~\ref{fig:grad_cam_O}, Fig.~\ref{fig:grad_cam_I}, Fig.~\ref{fig:grad_cam_M}, Fig.~\ref{fig:grad_cam_C}. The visualization shows that our framework places a primary emphasis on the internal regions of real faces, utilizing these areas as key indicators for classification. In contrast, when analyzing spoof images, the attention is more dispersed, concentrating on tell-tale signs of spoofing such as the edges of photos and features induced by screens. The results show the strong capability of \texttt{TeG-DG} to distinguish real and spoof features in unseen domains on the four datasets.

\begin{figure}[t]
\vspace{0.3cm}
\centering
\includegraphics[width=3.25in]{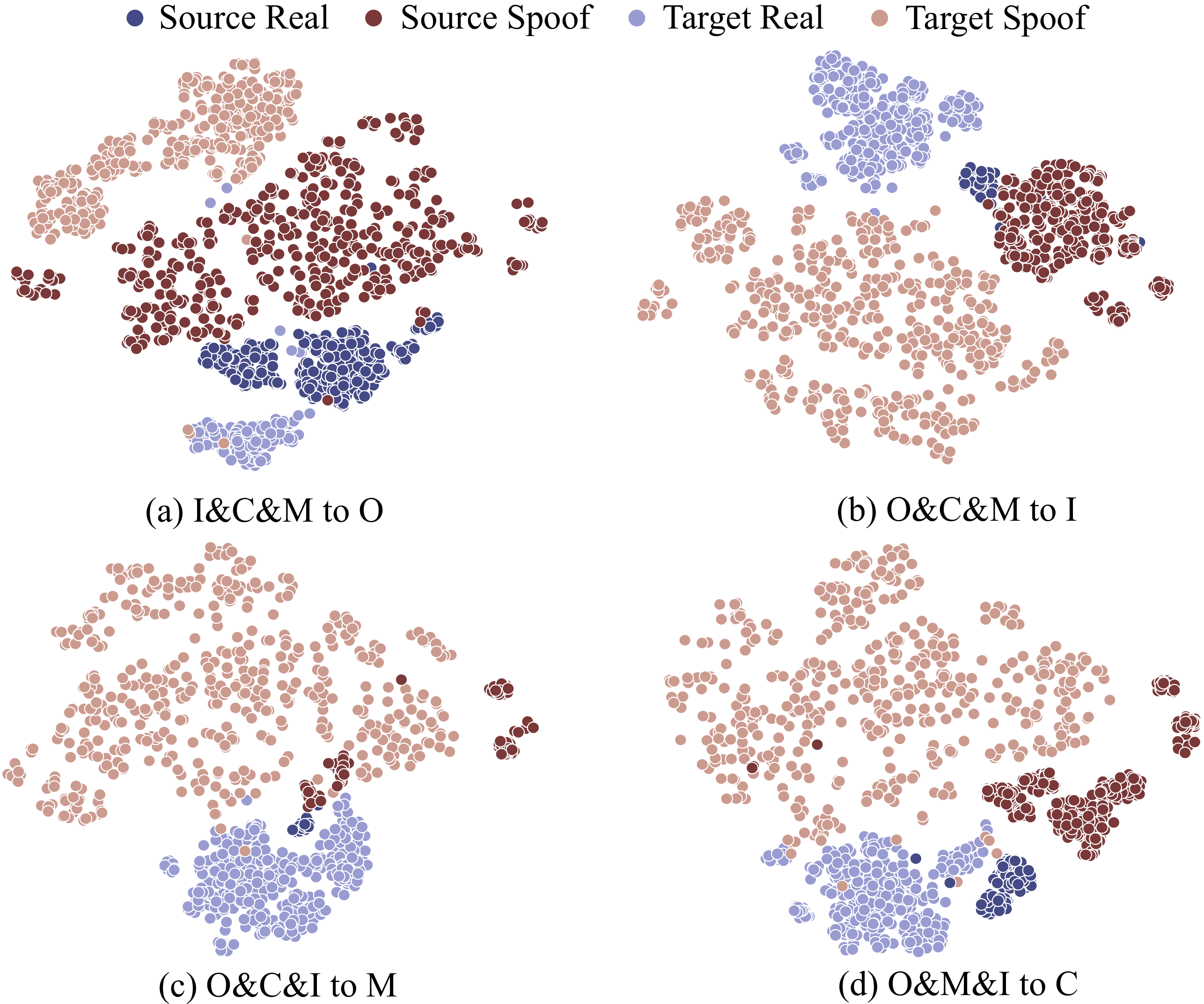}
\caption{\textbf{The t-SNE feature visualization on four LOO tests.}}
\label{fig_4dataset_tsne}
\end{figure}

We further elucidate the efficacy of the Textually Guided Domain Generalization (\texttt{TeG-DG}) framework through detailed t-SNE visualizations~\cite{van2008visualizing} across four LOO tests in Fig.~\ref{fig_4dataset_tsne}. These visualizations are instrumental in illustrating the practical impact of textual supervision on feature distribution and domain generalization. They facilitate a more pronounced separation in the feature space between real and spoof faces, while simultaneously achieving denser and more generalized clustering of real face features.

\section{Future Works}

In our future work, we aim to advance the field of Face Anti-Spoofing (FAS) through several innovative approaches. A key area of focus will be the development of methods for automatically generating appropriate text prompts for each training sample, rather than selecting them from the prompt library according to the type label of the image. This individualized approach to prompt generation is expected to enhance the effectiveness of our current framework.

Another critical aspect of our research will involve an in-depth investigation into the impact of various text prompts on training outcomes. We intend to identify the characteristics that make a text prompt effective or ineffective. Understanding the nuances of how different prompts influence training is essential for grasping the mechanics of prompt effectiveness in our system. Additionally, we plan to address the challenges of instability that arise from diverse combinations of text prompts.

Another pivotal aspect of our future work will involve integrating traditional domain generalization FAS techniques~\cite{SSDG_FAS, shao2019multi, SSAN_FAS, IADG_FAS, liao2023domain} with text-supervised methods. This integration is anticipated to further enhance the performance of our models, leveraging the strengths of both domain generalization and text-based approaches to create more robust and efficient FAS systems.

Overall, our future endeavors aim to push the boundaries of domain generalization in face anti-spoofing, using text as a powerful tool to achieve superior model performance and understanding ability.

\clearpage
\begin{figure*}[t]
    \centering
    \begin{subfigure}{0.12\textwidth}
        \centering
        \includegraphics[width=\linewidth]{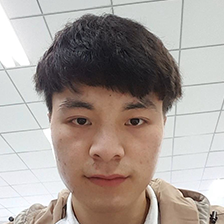}
    \end{subfigure}\hfil
    \begin{subfigure}{0.12\textwidth}
        \centering
        \includegraphics[width=\linewidth]{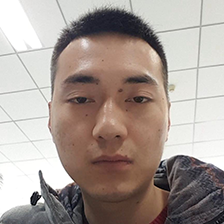}
    \end{subfigure}\hfil
    \begin{subfigure}{0.12\textwidth}
        \centering
        \includegraphics[width=\linewidth]{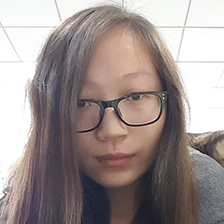}
    \end{subfigure}
    \hspace{8mm}
    \begin{subfigure}{0.12\textwidth}
        \centering
        \includegraphics[width=\linewidth]{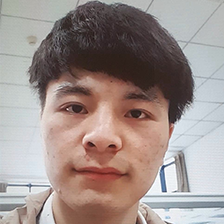}
    \end{subfigure}\hfil
    \begin{subfigure}{0.12\textwidth}
        \centering
        \includegraphics[width=\linewidth]{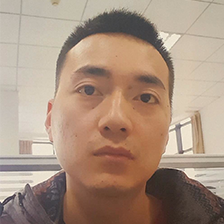}
    \end{subfigure}\hfil
    \begin{subfigure}{0.12\textwidth}
        \centering
        \includegraphics[width=\linewidth]{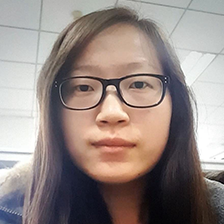}
    \end{subfigure}

    \vspace{5mm}

    \centering
    \begin{subfigure}{0.12\textwidth}
        \centering
        \includegraphics[width=\linewidth]{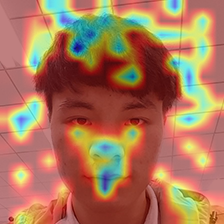}
    \end{subfigure}\hfil
    \begin{subfigure}{0.12\textwidth}
        \centering
        \includegraphics[width=\linewidth]{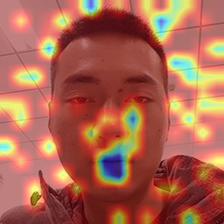}
    \end{subfigure}\hfil
    \begin{subfigure}{0.12\textwidth}
        \centering
        \includegraphics[width=\linewidth]{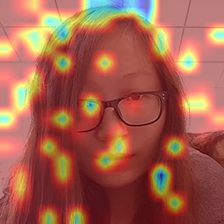}
    \end{subfigure}
    \hspace{8mm}
    \begin{subfigure}{0.12\textwidth}
        \centering
        \includegraphics[width=\linewidth]{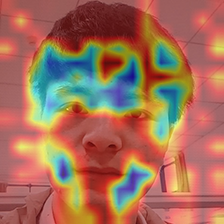}
    \end{subfigure}\hfil
    \begin{subfigure}{0.12\textwidth}
        \centering
        \includegraphics[width=\linewidth]{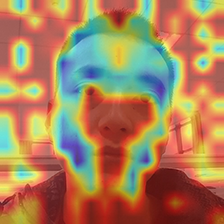}
    \end{subfigure}\hfil
    \begin{subfigure}{0.12\textwidth}
        \centering
        \includegraphics[width=\linewidth]{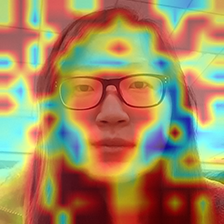}
    \end{subfigure}

    \vspace{-1mm}
    \caption{\textbf{Grad-Cam visualization for I\&C\&M to O.} On the left is a real image, and on the right is a spoof image.}
    \label{fig:grad_cam_O}
\end{figure*}

\vspace{-2.1cm}

\begin{figure*}[t]
    \centering
    \begin{subfigure}{0.12\textwidth}
        \centering
        \includegraphics[width=\linewidth]{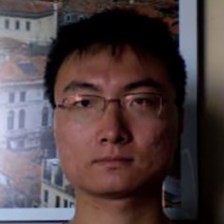}
    \end{subfigure}\hfil
    \begin{subfigure}{0.12\textwidth}
        \centering
        \includegraphics[width=\linewidth]{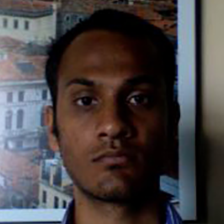}
    \end{subfigure}\hfil
    \begin{subfigure}{0.12\textwidth}
        \centering
        \includegraphics[width=\linewidth]{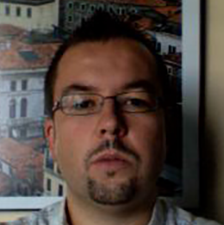}
    \end{subfigure}
    \hspace{8mm}
    \begin{subfigure}{0.12\textwidth}
        \centering
        \includegraphics[width=\linewidth]{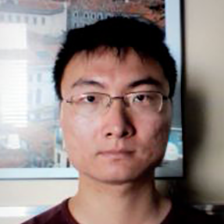}
    \end{subfigure}\hfil
    \begin{subfigure}{0.12\textwidth}
        \centering
        \includegraphics[width=\linewidth]{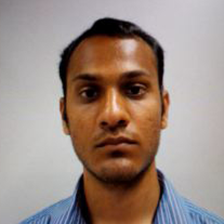}
    \end{subfigure}\hfil
    \begin{subfigure}{0.12\textwidth}
        \centering
        \includegraphics[width=\linewidth]{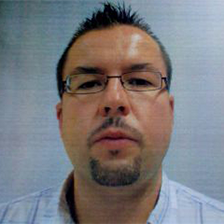}
    \end{subfigure}

    \vspace{5mm}
    
    \centering
    \begin{subfigure}{0.12\textwidth}
        \centering
        \includegraphics[width=\linewidth]{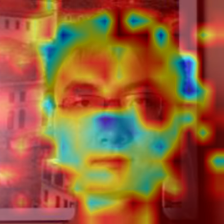}
    \end{subfigure}\hfil
    \begin{subfigure}{0.12\textwidth}
        \centering
        \includegraphics[width=\linewidth]{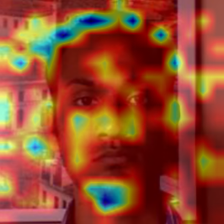}
    \end{subfigure}\hfil
    \begin{subfigure}{0.12\textwidth}
        \centering
        \includegraphics[width=\linewidth]{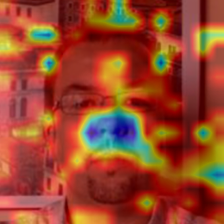}
    \end{subfigure}
    \hspace{8mm}
    \begin{subfigure}{0.12\textwidth}
        \centering
        \includegraphics[width=\linewidth]{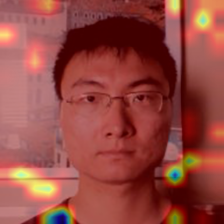}
    \end{subfigure}\hfil
    \begin{subfigure}{0.12\textwidth}
        \centering
        \includegraphics[width=\linewidth]{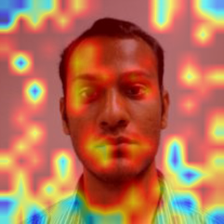}
    \end{subfigure}\hfil
    \begin{subfigure}{0.12\textwidth}
        \centering
        \includegraphics[width=\linewidth]{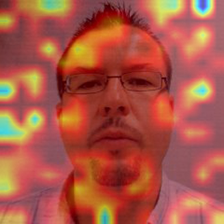}
    \end{subfigure}

    \vspace{-1mm}
    \caption{\textbf{Grad-Cam visualization for O\&C\&M to I.} On the left is a real image, and on the right is a spoof image.}
    \label{fig:grad_cam_I}
\end{figure*}

\vspace{-2.1cm}

\begin{figure*}[t]
    \centering
    \begin{subfigure}{0.12\textwidth}
        \centering
        \includegraphics[width=\linewidth]{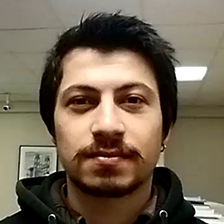}
    \end{subfigure}\hfil
    \begin{subfigure}{0.12\textwidth}
        \centering
        \includegraphics[width=\linewidth]{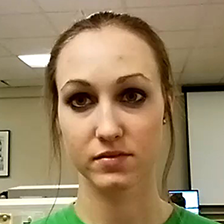}
    \end{subfigure}\hfil
    \begin{subfigure}{0.12\textwidth}
        \centering
        \includegraphics[width=\linewidth]{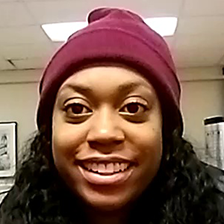}
    \end{subfigure}
    \hspace{8mm}
    \begin{subfigure}{0.12\textwidth}
        \centering
        \includegraphics[width=\linewidth]{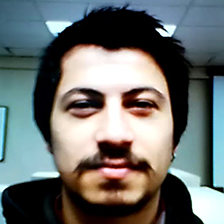}
    \end{subfigure}\hfil
    \begin{subfigure}{0.12\textwidth}
        \centering
        \includegraphics[width=\linewidth]{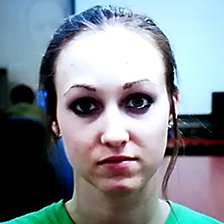}
    \end{subfigure}\hfil
    \begin{subfigure}{0.12\textwidth}
        \centering
        \includegraphics[width=\linewidth]{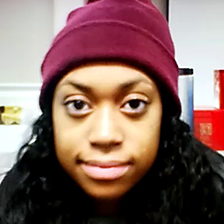}
    \end{subfigure}

    \vspace{5mm}

    \centering
    \begin{subfigure}{0.12\textwidth}
        \centering
        \includegraphics[width=\linewidth]{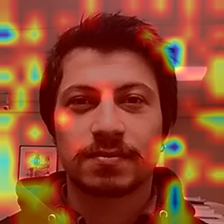}
    \end{subfigure}\hfil
    \begin{subfigure}{0.12\textwidth}
        \centering
        \includegraphics[width=\linewidth]{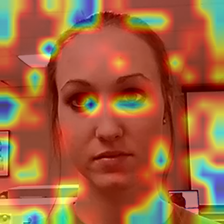}
    \end{subfigure}\hfil
    \begin{subfigure}{0.12\textwidth}
        \centering
        \includegraphics[width=\linewidth]{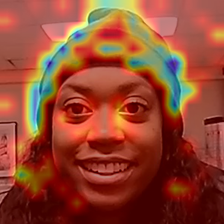}
    \end{subfigure}
    \hspace{8mm}
    \begin{subfigure}{0.12\textwidth}
        \centering
        \includegraphics[width=\linewidth]{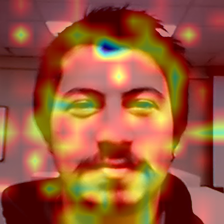}
    \end{subfigure}\hfil
    \begin{subfigure}{0.12\textwidth}
        \centering
        \includegraphics[width=\linewidth]{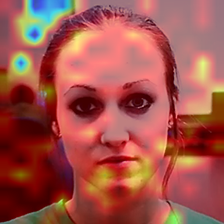}
    \end{subfigure}\hfil
    \begin{subfigure}{0.12\textwidth}
        \centering
        \includegraphics[width=\linewidth]{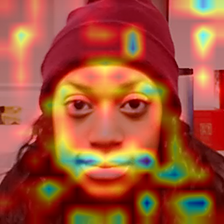}
    \end{subfigure}

    \vspace{-1mm}
    \caption{\textbf{Grad-Cam visualization for O\&C\&I to M.} On the left is a real image, and on the right is a spoof image.}
    \label{fig:grad_cam_M}
\end{figure*}

\vspace{-2.1cm}

\begin{figure*}[t]
    \centering
    \begin{subfigure}{0.12\textwidth}
        \centering
        \includegraphics[width=\linewidth]{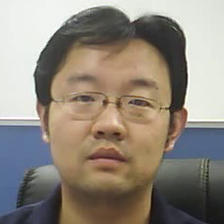}
    \end{subfigure}\hfil
    \begin{subfigure}{0.12\textwidth}
        \centering
        \includegraphics[width=\linewidth]{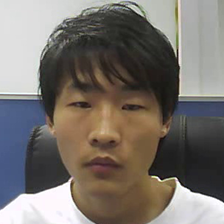}
    \end{subfigure}\hfil
    \begin{subfigure}{0.12\textwidth}
        \centering
        \includegraphics[width=\linewidth]{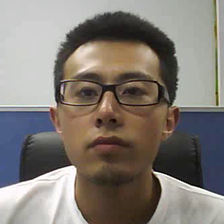}
    \end{subfigure}
    \hspace{8mm}
    \begin{subfigure}{0.12\textwidth}
        \centering
        \includegraphics[width=\linewidth]{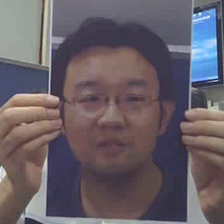}
    \end{subfigure}\hfil
    \begin{subfigure}{0.12\textwidth}
        \centering
        \includegraphics[width=\linewidth]{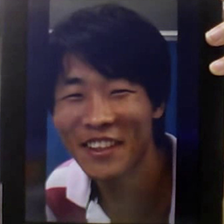}
    \end{subfigure}\hfil
    \begin{subfigure}{0.12\textwidth}
        \centering
        \includegraphics[width=\linewidth]{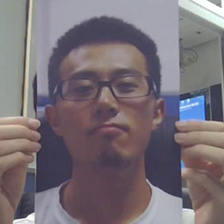}
    \end{subfigure}

    \vspace{5mm}

    \centering
    \begin{subfigure}{0.12\textwidth}
        \centering
        \includegraphics[width=\linewidth]{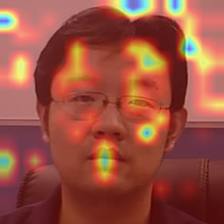}
    \end{subfigure}\hfil
    \begin{subfigure}{0.12\textwidth}
        \centering
        \includegraphics[width=\linewidth]{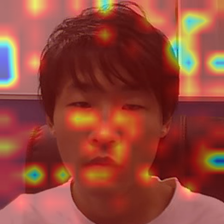}
    \end{subfigure}\hfil
    \begin{subfigure}{0.12\textwidth}
        \centering
        \includegraphics[width=\linewidth]{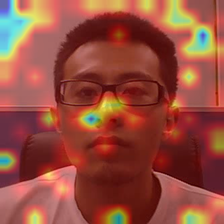}
    \end{subfigure}
    \hspace{8mm}
    \begin{subfigure}{0.12\textwidth}
        \centering
        \includegraphics[width=\linewidth]{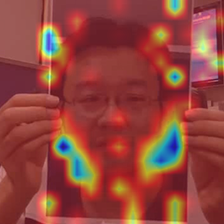}
    \end{subfigure}\hfil
    \begin{subfigure}{0.12\textwidth}
        \centering
        \includegraphics[width=\linewidth]{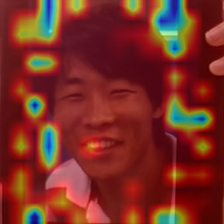}
    \end{subfigure}\hfil
    \begin{subfigure}{0.12\textwidth}
        \centering
        \includegraphics[width=\linewidth]{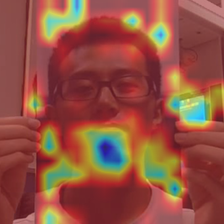}
    \end{subfigure}

    \vspace{-1mm}
    \caption{\textbf{Grad-Cam visualization for O\&M\&I to C.} On the left is a real image, and on the right is a spoof image.}
    \label{fig:grad_cam_C}
\end{figure*}

\clearpage
\begin{figure*}[t]
    \centering
    \begin{subfigure}{0.5\textwidth}
        \centering
        \includegraphics[width=\textwidth]{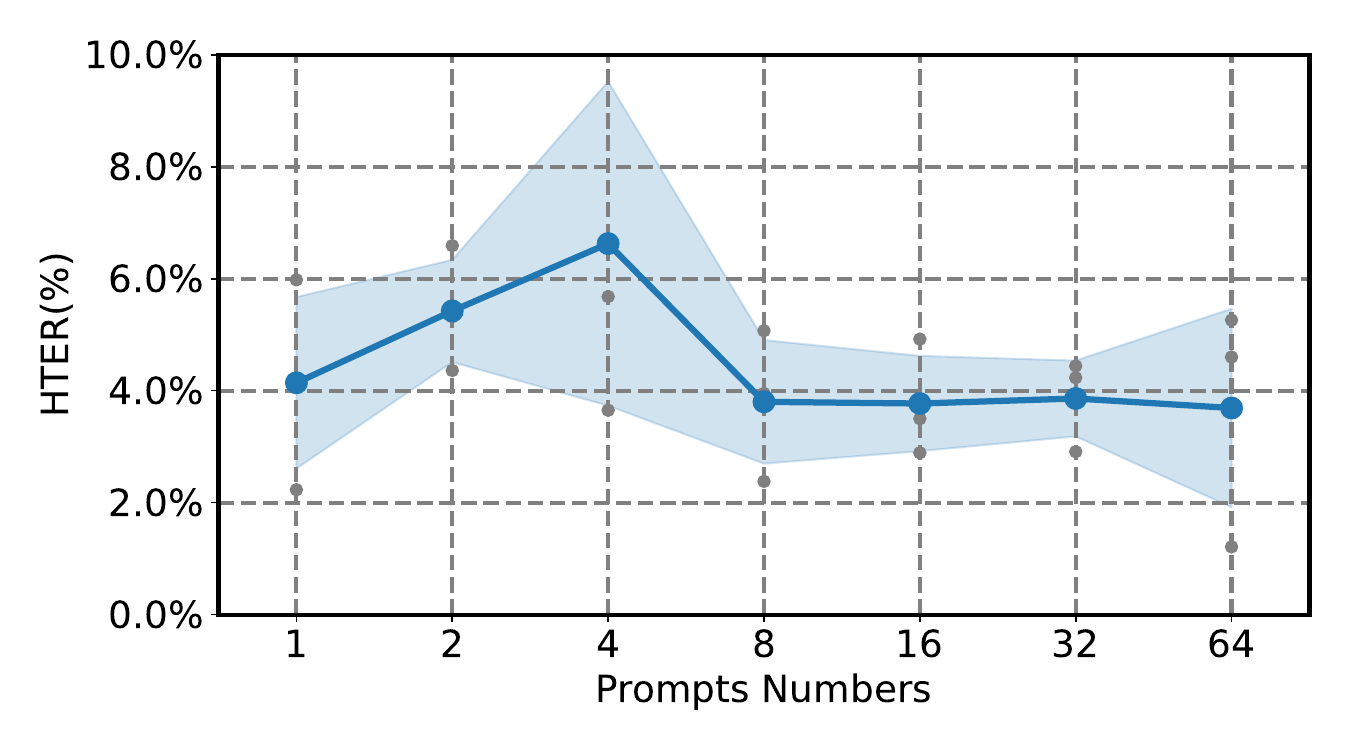}
    \end{subfigure}\hfill
    \begin{subfigure}{0.5\textwidth}
        \centering
        \includegraphics[width=\textwidth]{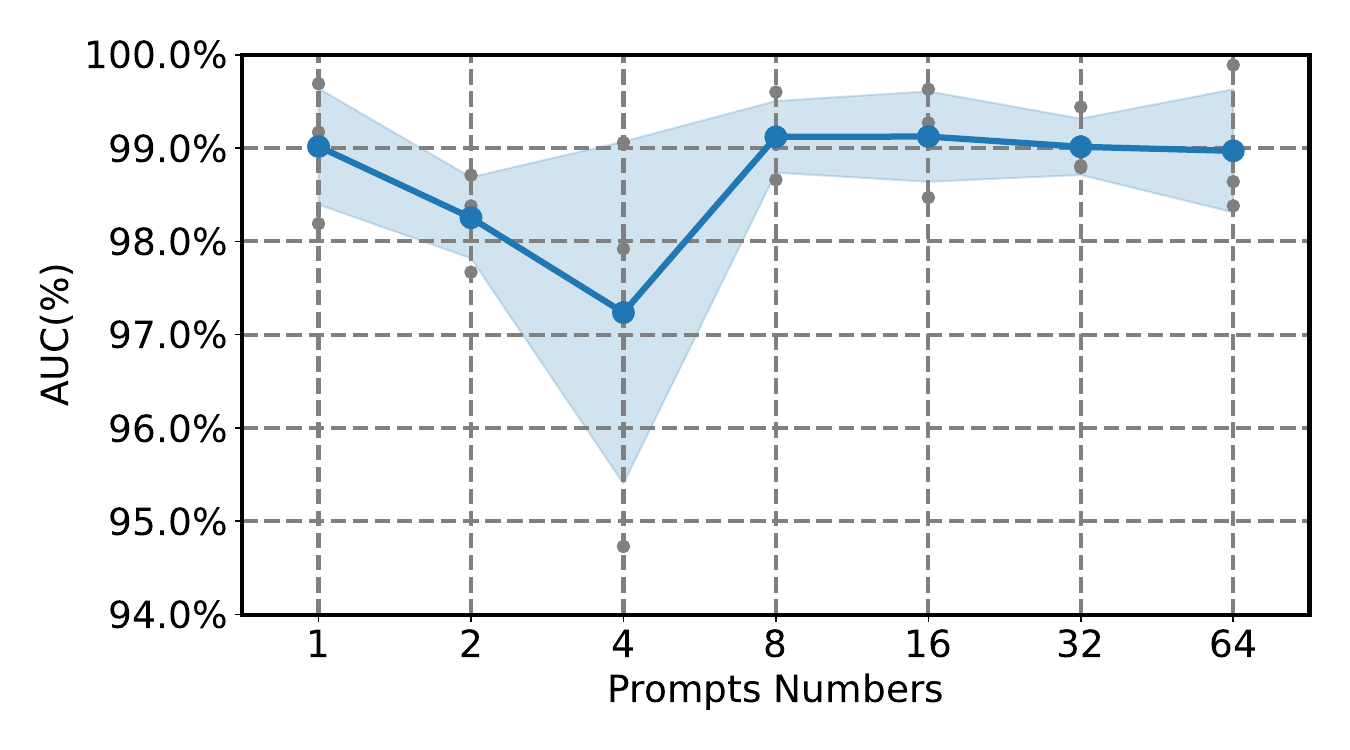}
    \end{subfigure}
    \vspace{-0.8cm}
    \caption{\textbf{HTER and AUC on I\&C\&M to O} under different text prompts number}
    \label{fig:prompt_ICM_O}
\end{figure*}

\begin{figure*}[t]
    \centering
    \begin{subfigure}{0.5\textwidth}
        \centering
        \includegraphics[width=\textwidth]{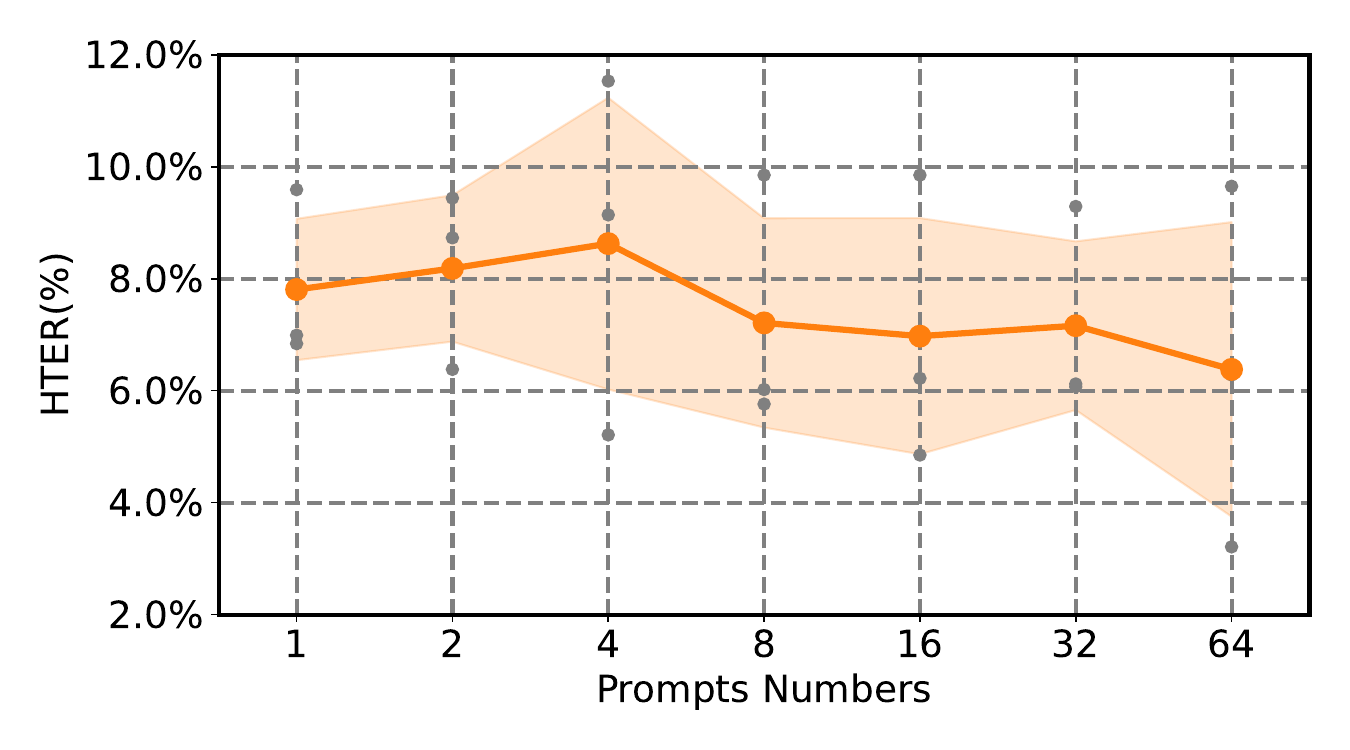}
    \end{subfigure}\hfill
    \begin{subfigure}{0.5\textwidth}
        \centering
        \includegraphics[width=\textwidth]{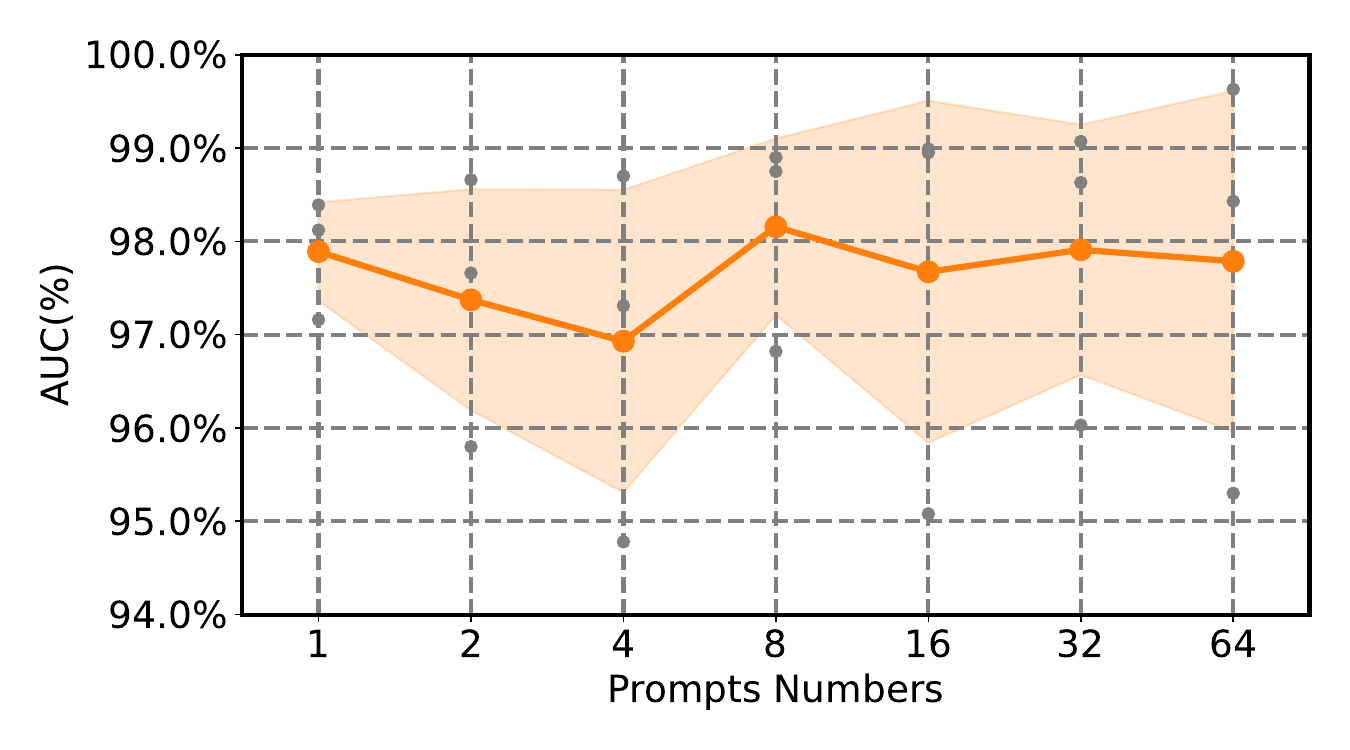}
    \end{subfigure}
    \vspace{-0.8cm}
    \caption{\textbf{HTER and AUC on O\&C\&M to I} under different text prompts number}
    \label{fig:prompt_OCM_I}
\end{figure*}

\begin{figure*}[t]
    \centering
    \begin{subfigure}{0.5\textwidth}
        \centering
        \includegraphics[width=\textwidth]{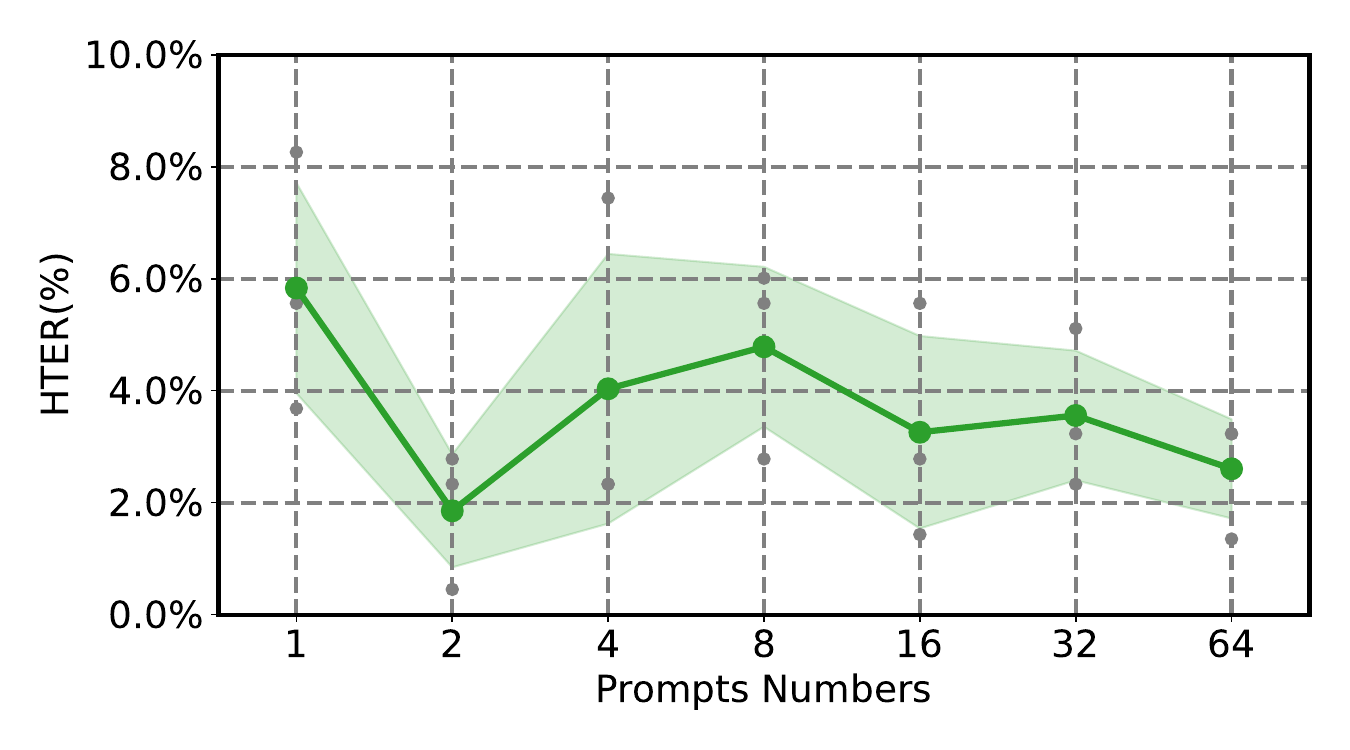}
    \end{subfigure}\hfill
    \begin{subfigure}{0.5\textwidth}
        \centering
        \includegraphics[width=\textwidth]{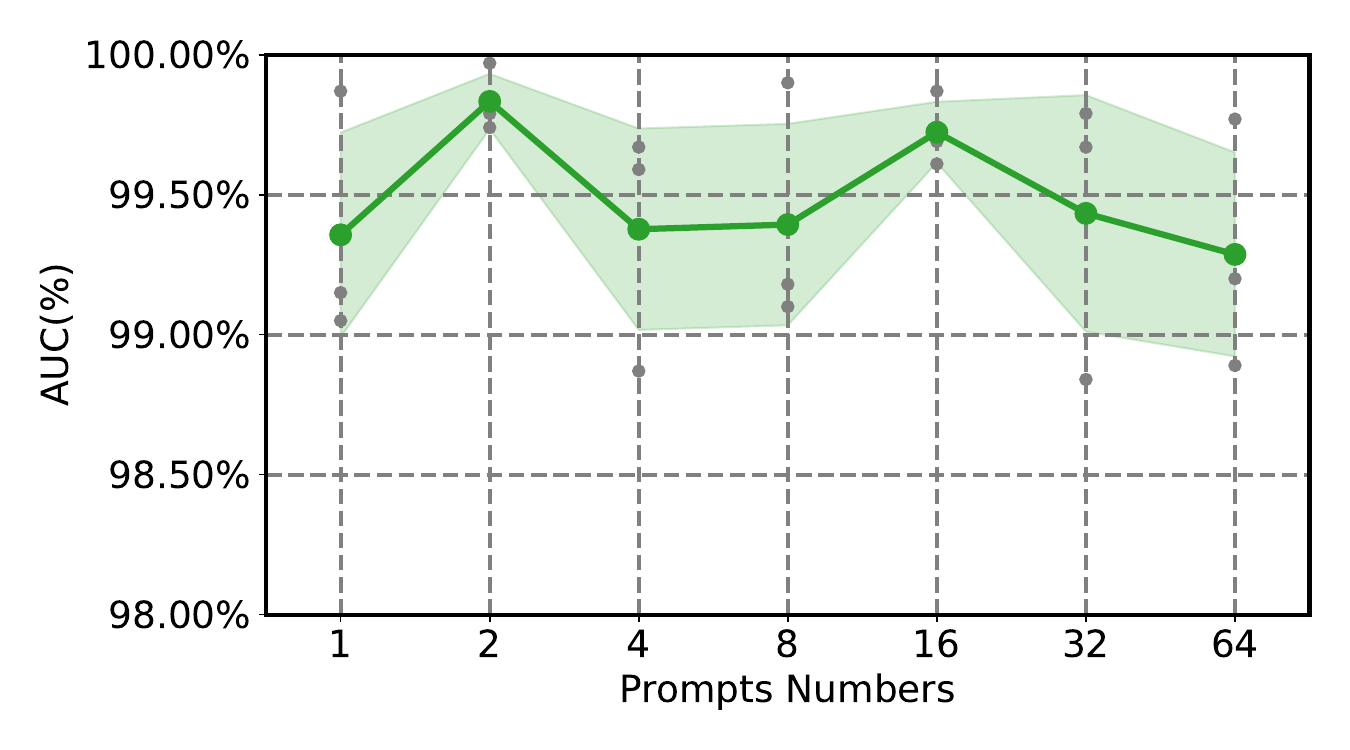}
    \end{subfigure}
    \vspace{-0.8cm}
    \caption{\textbf{HTER and AUC on O\&C\&I to M} under different text prompts number}
    \label{fig:prompt_OCI_M}
\end{figure*}

\begin{figure*}[t]
    \centering
    \begin{subfigure}{0.5\textwidth}
        \centering
        \includegraphics[width=\textwidth]{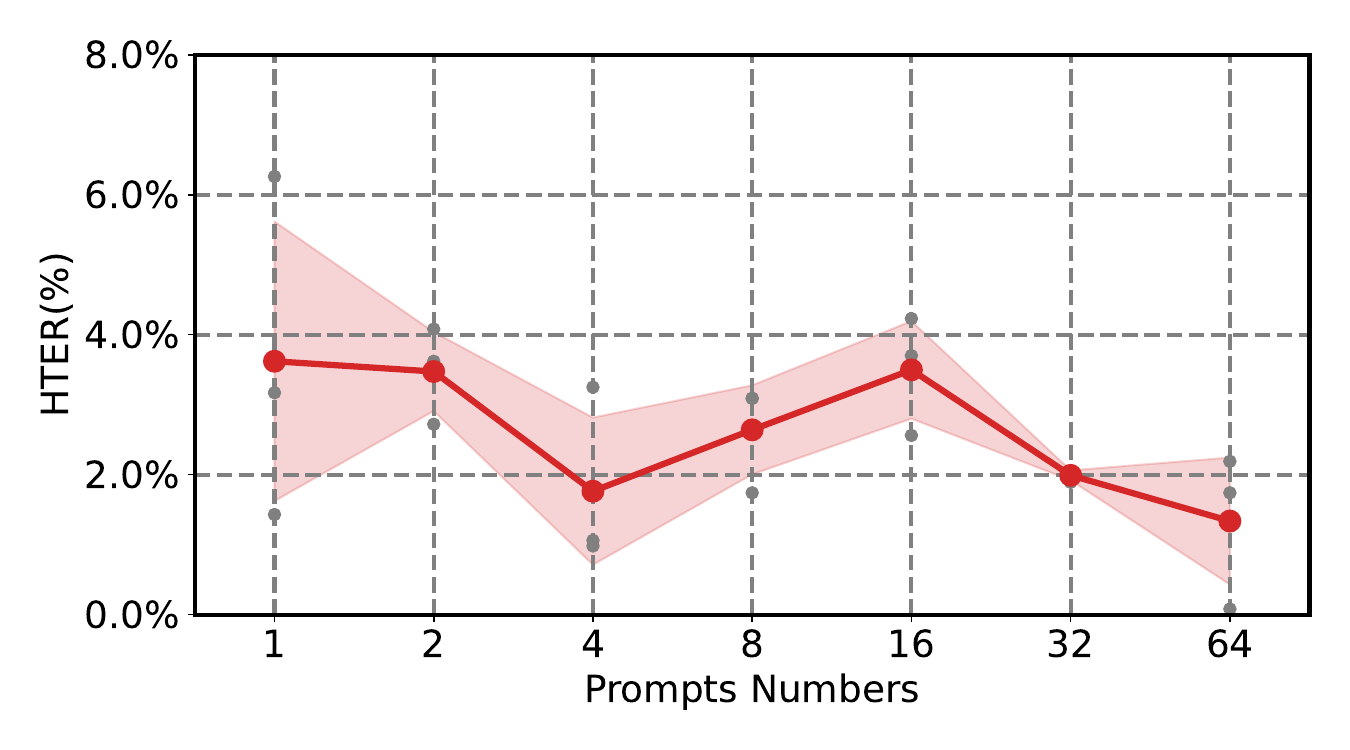}
    \end{subfigure}\hfill
    \begin{subfigure}{0.5\textwidth}
        \centering
        \includegraphics[width=\textwidth]{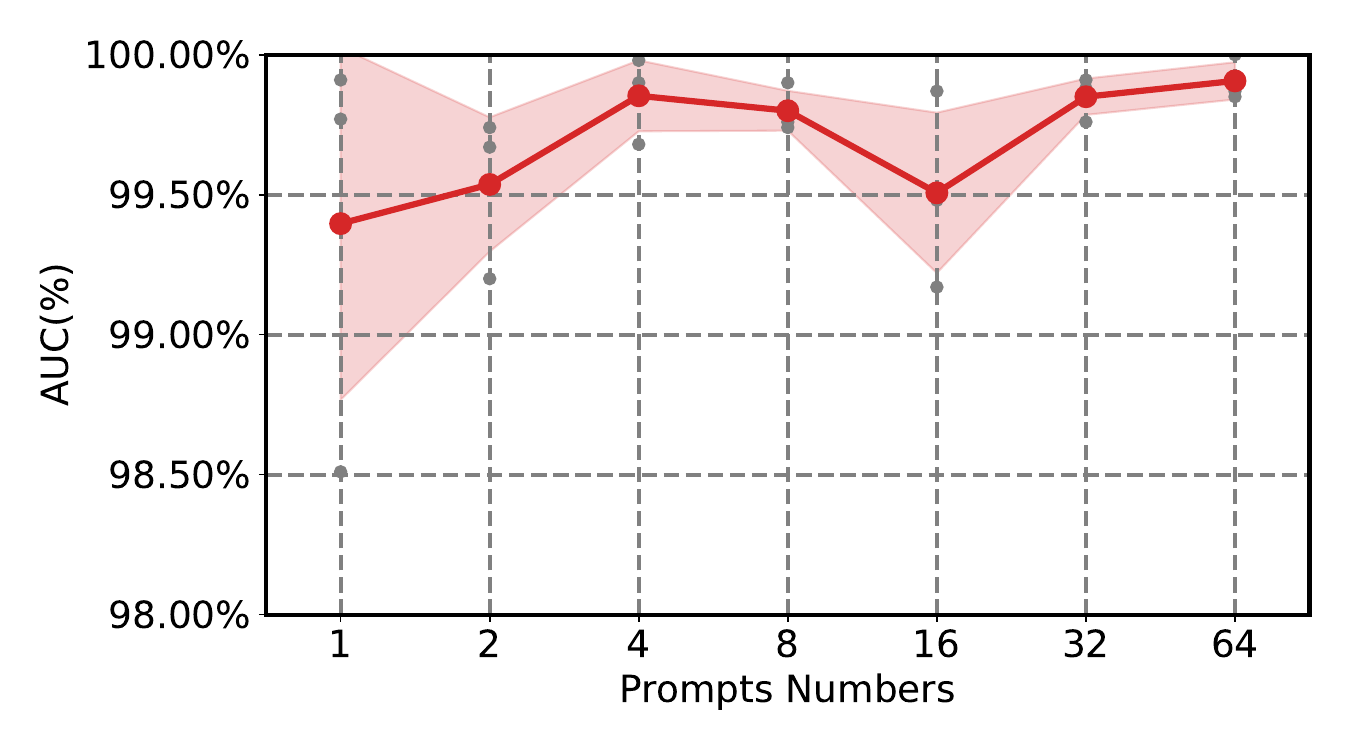}
    \end{subfigure}
    \vspace{-0.8cm}
    \caption{\textbf{HTER and AUC on O\&M\&I to C} under different text prompts number}
    \label{fig:prompt_OMI_C}
\end{figure*}

\clearpage
\clearpage
\section{Detailed Text Prompts}
\label{text_prompts}
\noindent\rule{\linewidth}{1pt}
\textbf{Text Prompts For Real Faces In Prompt Library}
\begin{enumerate}
    \item \setlength{\fboxsep}{1pt}\colorbox{gray!20}{\parbox{\dimexpr\linewidth-2\fboxsep\relax}{A real photo of a person}}
    \item A genuine image of the person
    \item \setlength{\fboxsep}{1pt}\colorbox{gray!20}{\parbox{\dimexpr\linewidth-2\fboxsep\relax}{An actual snapshot of the individual}}
    \item A real-life photograph of the person
    \item \setlength{\fboxsep}{1pt}\colorbox{gray!20}{\parbox{\dimexpr\linewidth-2\fboxsep\relax}{A true-to-life photo of the person}}
    \item An authentic photograph of the individual
    \item \setlength{\fboxsep}{1pt}\colorbox{gray!20}{\parbox{\dimexpr\linewidth-2\fboxsep\relax}{A bona fide picture of the person}}
    \item An unedited photograph of the individual
    \item \setlength{\fboxsep}{1pt}\colorbox{gray!20}{\parbox{\dimexpr\linewidth-2\fboxsep\relax}{A legitimate snapshot of the person}}
    \item An original photo depicting the individual
    \item \setlength{\fboxsep}{1pt}\colorbox{gray!20}{\parbox{\dimexpr\linewidth-2\fboxsep\relax}{A veritable image of the person}}
    \item An unaltered photograph of the individual
    \item \setlength{\fboxsep}{1pt}\colorbox{gray!20}{\parbox{\dimexpr\linewidth-2\fboxsep\relax}{A factual representation in a photo of the person}}
    \item An honest capture of the individual's likeness
    \item \setlength{\fboxsep}{1pt}\colorbox{gray!20}{\parbox{\dimexpr\linewidth-2\fboxsep\relax}{A straightforward photograph of the person}}
    \item An undistorted image of the individual
    \item \setlength{\fboxsep}{1pt}\colorbox{gray!20}{\parbox{\dimexpr\linewidth-2\fboxsep\relax}{An unembellished photograph of the individual}}
    \item A straightforward capture of the person's appearance
    \item \setlength{\fboxsep}{1pt}\colorbox{gray!20}{\parbox{\dimexpr\linewidth-2\fboxsep\relax}{A direct shot of the individual, unmodified}}
    \item A candid photograph depicting the person
    \item \setlength{\fboxsep}{1pt}\colorbox{gray!20}{\parbox{\dimexpr\linewidth-2\fboxsep\relax}{A natural, unposed photo of the individual}}
    \item An unretouched image of the person
    \item \setlength{\fboxsep}{1pt}\colorbox{gray!20}{\parbox{\dimexpr\linewidth-2\fboxsep\relax}{A clear, unfiltered photograph of the individual}}
    \item A genuine representation in a photo of the person
    \item \setlength{\fboxsep}{1pt}\colorbox{gray!20}{\parbox{\dimexpr\linewidth-2\fboxsep\relax}{An honest, unadorned picture of the individual}}
    \item A pure, unedited image of the person
    \item \setlength{\fboxsep}{1pt}\colorbox{gray!20}{\parbox{\dimexpr\linewidth-2\fboxsep\relax}{A real, unmanipulated photograph of the individual}}
    \item A sincere, untouched snapshot of the person
    \item \setlength{\fboxsep}{1pt}\colorbox{gray!20}{\parbox{\dimexpr\linewidth-2\fboxsep\relax}{An exact, unaltered capture of the individual's visage}}
    \item An unvarnished, true photo of the person
    \item \setlength{\fboxsep}{1pt}\colorbox{gray!20}{\parbox{\dimexpr\linewidth-2\fboxsep\relax}{An unexaggerated, straightforward image of the individual}}
    \item A plain, accurate photograph of the person
    \item \setlength{\fboxsep}{1pt}\colorbox{gray!20}{\parbox{\dimexpr\linewidth-2\fboxsep\relax}{A genuine, unaltered image of the person}}
    \item A truthful depiction in a photograph of the individual
    \item \setlength{\fboxsep}{1pt}\colorbox{gray!20}{\parbox{\dimexpr\linewidth-2\fboxsep\relax}{An unprocessed photo of the person}}
    \item A raw, natural shot of the individual
    \item \setlength{\fboxsep}{1pt}\colorbox{gray!20}{\parbox{\dimexpr\linewidth-2\fboxsep\relax}{A direct, unaltered photograph of the person}}
    \item A true, unfiltered image of the individual
    \item \setlength{\fboxsep}{1pt}\colorbox{gray!20}{\parbox{\dimexpr\linewidth-2\fboxsep\relax}{An unmanipulated representation in a photo of the person}}
    \item A straightforward, unedited picture of the individual
    \item \setlength{\fboxsep}{1pt}\colorbox{gray!20}{\parbox{\dimexpr\linewidth-2\fboxsep\relax}{An undisturbed, authentic photo of the person}}
    \item An original, unenhanced image of the individual
    \item \setlength{\fboxsep}{1pt}\colorbox{gray!20}{\parbox{\dimexpr\linewidth-2\fboxsep\relax}{A non-doctored photograph of the person}}
    \item A pure, straightforward shot of the individual
    \item \setlength{\fboxsep}{1pt}\colorbox{gray!20}{\parbox{\dimexpr\linewidth-2\fboxsep\relax}{An unsullied, genuine photograph of the person}}
    \item An uncropped, clear image of the individual
    \item \setlength{\fboxsep}{1pt}\colorbox{gray!20}{\parbox{\dimexpr\linewidth-2\fboxsep\relax}{A veracious, unaltered photo of the person}}
    \item A true representation in a snapshot of the individual
    \item \setlength{\fboxsep}{1pt}\colorbox{gray!20}{\parbox{\dimexpr\linewidth-2\fboxsep\relax}{An honest, unfiltered view of the person}}
    \item A direct, unmanipulated photo of the individual
    \item \setlength{\fboxsep}{1pt}\colorbox{gray!20}{\parbox{\dimexpr\linewidth-2\fboxsep\relax}{A natural, unaltered image of the person}}
    \item An untouched, true-to-form photograph of the individual
    \item \setlength{\fboxsep}{1pt}\colorbox{gray!20}{\parbox{\dimexpr\linewidth-2\fboxsep\relax}{A straightforward, unenhanced snapshot of the person}}
    \item An unedited, authentic image of the individual
    \item \setlength{\fboxsep}{1pt}\colorbox{gray!20}{\parbox{\dimexpr\linewidth-2\fboxsep\relax}{A clear-cut, unadulterated photo of the person}}
    \item A non-altered, genuine representation of the individual
    \item \setlength{\fboxsep}{1pt}\colorbox{gray!20}{\parbox{\dimexpr\linewidth-2\fboxsep\relax}{A raw, unfiltered capture of the person}}
    \item An unadorned, straightforward picture of the individual
    \item \setlength{\fboxsep}{1pt}\colorbox{gray!20}{\parbox{\dimexpr\linewidth-2\fboxsep\relax}{A pure, unvarnished image of the person}}
    \item An honest, unprocessed photograph of the individual
    \item \setlength{\fboxsep}{1pt}\colorbox{gray!20}{\parbox{\dimexpr\linewidth-2\fboxsep\relax}{A direct, undistorted snapshot of the person}}
    \item An unedited, clear depiction of the individual
    \item \setlength{\fboxsep}{1pt}\colorbox{gray!20}{\parbox{\dimexpr\linewidth-2\fboxsep\relax}{A truthful, unaltered photo of the person}}
    \item An unenhanced, natural picture of the individual
\end{enumerate}
\noindent\rule{\linewidth}{1pt}

\clearpage
\noindent\rule{\linewidth}{1pt}
\textbf{Text Prompts For Print Attacks In Prompt Library}
\begin{enumerate}
    \item \setlength{\fboxsep}{1pt}\colorbox{gray!20}{\parbox{\dimexpr\linewidth-2\fboxsep\relax}{A printed photo}}
    \item A print attack photo
    \item \setlength{\fboxsep}{1pt}\colorbox{gray!20}{\parbox{\dimexpr\linewidth-2\fboxsep\relax}{A printed photo with is blur and lack of details}}
    \item A photo of an A4 paper with a face printed on it
    \item \setlength{\fboxsep}{1pt}\colorbox{gray!20}{\parbox{\dimexpr\linewidth-2\fboxsep\relax}{An image of an A4 sheet of paper bearing a printed face}}
    \item A photo of a paper printed with an image of a person
    \item \setlength{\fboxsep}{1pt}\colorbox{gray!20}{\parbox{\dimexpr\linewidth-2\fboxsep\relax}{A hard copy of a photograph}}
    \item A printed image on paper
    \item \setlength{\fboxsep}{1pt}\colorbox{gray!20}{\parbox{\dimexpr\linewidth-2\fboxsep\relax}{A blurred and indistinct printed photo}}
    \item A photograph depicting a face on an A4 sheet
    \item \setlength{\fboxsep}{1pt}\colorbox{gray!20}{\parbox{\dimexpr\linewidth-2\fboxsep\relax}{An A4 paper with a facial image printed upon it}}
    \item A paper bearing a printed photograph of an individual
    \item \setlength{\fboxsep}{1pt}\colorbox{gray!20}{\parbox{\dimexpr\linewidth-2\fboxsep\relax}{A printout of a photo with reduced clarity and detail}}
    \item An image of a person printed on standard paper
    \item \setlength{\fboxsep}{1pt}\colorbox{gray!20}{\parbox{\dimexpr\linewidth-2\fboxsep\relax}{A photo print, lacking sharpness and detail, on paper}}
    \item A snapshot printed on an A4-sized paper with a facial image
    \item \setlength{\fboxsep}{1pt}\colorbox{gray!20}{\parbox{\dimexpr\linewidth-2\fboxsep\relax}{A low-resolution printout of a face}}
    \item An image on paper, showing signs of being printed
    \item \setlength{\fboxsep}{1pt}\colorbox{gray!20}{\parbox{\dimexpr\linewidth-2\fboxsep\relax}{A photocopy of a photographic image}}
    \item A print of a digital photo, with visible pixelation
    \item \setlength{\fboxsep}{1pt}\colorbox{gray!20}{\parbox{\dimexpr\linewidth-2\fboxsep\relax}{A face image printed on a regular sheet of paper}}
    \item A printed photograph with faded colors
    \item \setlength{\fboxsep}{1pt}\colorbox{gray!20}{\parbox{\dimexpr\linewidth-2\fboxsep\relax}{A photo printout showing distortion}}
    \item A printed picture on a letter-sized paper
    \item \setlength{\fboxsep}{1pt}\colorbox{gray!20}{\parbox{\dimexpr\linewidth-2\fboxsep\relax}{An office printer output of a face photo}}
    \item A printed facial image with visible print lines
    \item \setlength{\fboxsep}{1pt}\colorbox{gray!20}{\parbox{\dimexpr\linewidth-2\fboxsep\relax}{A smudged photo printout on white paper}}
    \item A paper print of a digital image, showing compression artifacts
    \item \setlength{\fboxsep}{1pt}\colorbox{gray!20}{\parbox{\dimexpr\linewidth-2\fboxsep\relax}{A color print of a photograph, slightly blurred}}
    \item A printed image of a person with low contrast
    \item \setlength{\fboxsep}{1pt}\colorbox{gray!20}{\parbox{\dimexpr\linewidth-2\fboxsep\relax}{A printout of a portrait photo on plain paper}}
    \item A digitally printed photo with overexposure
    \item \setlength{\fboxsep}{1pt}\colorbox{gray!20}{\parbox{\dimexpr\linewidth-2\fboxsep\relax}{A facial photo printed using a home printer}}
    \item A printout of a photo, showing signs of ink spreading
    \item \setlength{\fboxsep}{1pt}\colorbox{gray!20}{\parbox{\dimexpr\linewidth-2\fboxsep\relax}{A printed image, lacking in fine details}}
    \item A photo on paper, with visible printing dots
    \item \setlength{\fboxsep}{1pt}\colorbox{gray!20}{\parbox{\dimexpr\linewidth-2\fboxsep\relax}{A printed portrait with a grainy texture}}
    \item A photo reproduction on matte paper
    \item \setlength{\fboxsep}{1pt}\colorbox{gray!20}{\parbox{\dimexpr\linewidth-2\fboxsep\relax}{A photocopy of a face, with some areas washed out}}
    \item A laser-printed image of an individual
    \item \setlength{\fboxsep}{1pt}\colorbox{gray!20}{\parbox{\dimexpr\linewidth-2\fboxsep\relax}{An inkjet printed photo with color misalignment}}
    \item A printed image, showing banding issues
    \item \setlength{\fboxsep}{1pt}\colorbox{gray!20}{\parbox{\dimexpr\linewidth-2\fboxsep\relax}{A paper print showing a digitally zoomed-in face}}
    \item A printed photo with noticeable color shifts
    \item \setlength{\fboxsep}{1pt}\colorbox{gray!20}{\parbox{\dimexpr\linewidth-2\fboxsep\relax}{A facial photograph printed on glossy paper}}
    \item A printout of a photo with a moiré pattern
    \item \setlength{\fboxsep}{1pt}\colorbox{gray!20}{\parbox{\dimexpr\linewidth-2\fboxsep\relax}{A printed face photo with a skewed perspective}}
    \item A color photo printed on a grayscale setting
    \item \setlength{\fboxsep}{1pt}\colorbox{gray!20}{\parbox{\dimexpr\linewidth-2\fboxsep\relax}{A photocopy of a person's photo, with reduced saturation}}
    \item A printed image of a face, slightly off-center on the paper
    \item \setlength{\fboxsep}{1pt}\colorbox{gray!20}{\parbox{\dimexpr\linewidth-2\fboxsep\relax}{A photo print showing ink smears}}
    \item A printed image with a noticeable paper texture
    \item \setlength{\fboxsep}{1pt}\colorbox{gray!20}{\parbox{\dimexpr\linewidth-2\fboxsep\relax}{A photograph print with uneven ink distribution}}
    \item A printed photo, showing reduced dynamic range
    \item \setlength{\fboxsep}{1pt}\colorbox{gray!20}{\parbox{\dimexpr\linewidth-2\fboxsep\relax}{A face printout on textured paper}}
    \item A photo print with a yellowish tint
    \item \setlength{\fboxsep}{1pt}\colorbox{gray!20}{\parbox{\dimexpr\linewidth-2\fboxsep\relax}{A printed image of a person, cropped awkwardly}}
    \item A facial photo printed with low ink levels
    \item \setlength{\fboxsep}{1pt}\colorbox{gray!20}{\parbox{\dimexpr\linewidth-2\fboxsep\relax}{A digitally printed face with artifacts}}
    \item A printout of a photo with a watermark
    \item \setlength{\fboxsep}{1pt}\colorbox{gray!20}{\parbox{\dimexpr\linewidth-2\fboxsep\relax}{A printed photograph, slightly torn at the edge}}
    \item A print of a digital photo, with color bleeding
    \item \setlength{\fboxsep}{1pt}\colorbox{gray!20}{\parbox{\dimexpr\linewidth-2\fboxsep\relax}{A photo printed on thin, low-quality paper}}
    \item A printout of a face, showing digital noise
\end{enumerate}
\noindent\rule{\linewidth}{1pt}

\clearpage
\noindent\rule{\linewidth}{1pt}
\textbf{Text Prompts For Replay Attacks In Prompt Library}
\begin{enumerate}
    \item \setlength{\fboxsep}{1pt}\colorbox{gray!20}{\parbox{\dimexpr\linewidth-2\fboxsep\relax}{A replay attack photo}}
    \item A photo of an iPad
    \item \setlength{\fboxsep}{1pt}\colorbox{gray!20}{\parbox{\dimexpr\linewidth-2\fboxsep\relax}{A photo of an iPad that displaying face images}}
    \item A photo of a display
    \item \setlength{\fboxsep}{1pt}\colorbox{gray!20}{\parbox{\dimexpr\linewidth-2\fboxsep\relax}{A photo of a screen}}
    \item An image of a screen
    \item \setlength{\fboxsep}{1pt}\colorbox{gray!20}{\parbox{\dimexpr\linewidth-2\fboxsep\relax}{A photograph used for a replay attack}}
    \item An image of an iPad
    \item \setlength{\fboxsep}{1pt}\colorbox{gray!20}{\parbox{\dimexpr\linewidth-2\fboxsep\relax}{A photo showing an iPad displaying facial images}}
    \item A snapshot of a digital display screen
    \item \setlength{\fboxsep}{1pt}\colorbox{gray!20}{\parbox{\dimexpr\linewidth-2\fboxsep\relax}{A photograph capturing a computer or television screen}}
    \item An image depicting a monitor display
    \item \setlength{\fboxsep}{1pt}\colorbox{gray!20}{\parbox{\dimexpr\linewidth-2\fboxsep\relax}{A picture of a tablet screen showing a face}}
    \item A photograph of a device's screen in operation
    \item \setlength{\fboxsep}{1pt}\colorbox{gray!20}{\parbox{\dimexpr\linewidth-2\fboxsep\relax}{An image depicting a monitor display}}
    \item A picture of a tablet screen showing a face
    \item \setlength{\fboxsep}{1pt}\colorbox{gray!20}{\parbox{\dimexpr\linewidth-2\fboxsep\relax}{A photo of a smartphone screen displaying an image}}
    \item An image of a laptop screen showing a face
    \item \setlength{\fboxsep}{1pt}\colorbox{gray!20}{\parbox{\dimexpr\linewidth-2\fboxsep\relax}{A snapshot showing a monitor with a facial image}}
    \item A photo of a device screen during a replay attack
    \item \setlength{\fboxsep}{1pt}\colorbox{gray!20}{\parbox{\dimexpr\linewidth-2\fboxsep\relax}{An image of a screen showing a video playback}}
    \item A photograph of a face displayed on a digital device
    \item \setlength{\fboxsep}{1pt}\colorbox{gray!20}{\parbox{\dimexpr\linewidth-2\fboxsep\relax}{A picture capturing a face on a tablet display}}
    \item A photo of a monitor screen with a facial portrait
    \item \setlength{\fboxsep}{1pt}\colorbox{gray!20}{\parbox{\dimexpr\linewidth-2\fboxsep\relax}{An image of a screen replaying a video}}
    \item A photograph showing a digital screen in use
    \item \setlength{\fboxsep}{1pt}\colorbox{gray!20}{\parbox{\dimexpr\linewidth-2\fboxsep\relax}{A snapshot of a face being displayed on a smartphone}}
    \item A photo of a computer monitor displaying images
    \item \setlength{\fboxsep}{1pt}\colorbox{gray!20}{\parbox{\dimexpr\linewidth-2\fboxsep\relax}{An image of a television screen in operation}}
    \item A picture of a digital display showing a live feed
    \item \setlength{\fboxsep}{1pt}\colorbox{gray!20}{\parbox{\dimexpr\linewidth-2\fboxsep\relax}{A photo of an electronic display with a static image}}
    \item An image capturing a tablet screen in operation
    \item \setlength{\fboxsep}{1pt}\colorbox{gray!20}{\parbox{\dimexpr\linewidth-2\fboxsep\relax}{A photograph of a screen showing a streaming video}}
    \item A snapshot of a digital billboard display
    \item \setlength{\fboxsep}{1pt}\colorbox{gray!20}{\parbox{\dimexpr\linewidth-2\fboxsep\relax}{A photo capturing a screen with multimedia content}}
    \item An image of a projected screen displaying a face
    \item \setlength{\fboxsep}{1pt}\colorbox{gray!20}{\parbox{\dimexpr\linewidth-2\fboxsep\relax}{A picture of a face on a high-resolution monitor}}
    \item A photo of a screen with high brightness showing a face
    \item \setlength{\fboxsep}{1pt}\colorbox{gray!20}{\parbox{\dimexpr\linewidth-2\fboxsep\relax}{An image of a LED display screen in action}}
    \item A photograph of a smartphone screen with a zoomed-in face
    \item \setlength{\fboxsep}{1pt}\colorbox{gray!20}{\parbox{\dimexpr\linewidth-2\fboxsep\relax}{A picture showing a monitor with a live webcam feed}}
    \item A snapshot of a TV screen showing a recorded video
    \item \setlength{\fboxsep}{1pt}\colorbox{gray!20}{\parbox{\dimexpr\linewidth-2\fboxsep\relax}{A photo of a video call in progress on a tablet}}
    \item An image showing a digital kiosk display
    \item \setlength{\fboxsep}{1pt}\colorbox{gray!20}{\parbox{\dimexpr\linewidth-2\fboxsep\relax}{A photograph of a computer screen with a face slideshow}}
    \item A picture of a mobile phone screen playing a video
    \item \setlength{\fboxsep}{1pt}\colorbox{gray!20}{\parbox{\dimexpr\linewidth-2\fboxsep\relax}{A photo showing a digital photo frame in use}}
    \item An image of a virtual reality screen displaying a face
    \item \setlength{\fboxsep}{1pt}\colorbox{gray!20}{\parbox{\dimexpr\linewidth-2\fboxsep\relax}{A picture showing a computer monitor with editing software}}
    \item A snapshot of an outdoor LED screen displaying an ad
    \item \setlength{\fboxsep}{1pt}\colorbox{gray!20}{\parbox{\dimexpr\linewidth-2\fboxsep\relax}{A photo of a tablet screen with a live streaming app}}
    \item An image of a monitor with a split-screen view
    \item \setlength{\fboxsep}{1pt}\colorbox{gray!20}{\parbox{\dimexpr\linewidth-2\fboxsep\relax}{A photograph of a phone screen showing a photo gallery}}
    \item A picture of a screen displaying interactive media
    \item \setlength{\fboxsep}{1pt}\colorbox{gray!20}{\parbox{\dimexpr\linewidth-2\fboxsep\relax}{A photo showing a touchscreen device in use}}
    \item An image of a monitor with screen mirroring active
    \item \setlength{\fboxsep}{1pt}\colorbox{gray!20}{\parbox{\dimexpr\linewidth-2\fboxsep\relax}{A photograph showing a tablet with a drawing app}}
    \item A picture of a laptop screen with a video conference
    \item \setlength{\fboxsep}{1pt}\colorbox{gray!20}{\parbox{\dimexpr\linewidth-2\fboxsep\relax}{A snapshot of a smartwatch screen with a notification}}
    \item A photo of a car display screen showing navigation
    \item \setlength{\fboxsep}{1pt}\colorbox{gray!20}{\parbox{\dimexpr\linewidth-2\fboxsep\relax}{An image of a screen with augmented reality content}}
    \item A photograph of a display with facial recognition software
    \item \setlength{\fboxsep}{1pt}\colorbox{gray!20}{\parbox{\dimexpr\linewidth-2\fboxsep\relax}{A picture of a screen displaying a 3D model of a face}}
    \item A photo of a digital signboard showing a portrait
\end{enumerate}

\noindent\rule{\linewidth}{1pt} \fi

\clearpage
{\small
\bibliographystyle{ieeenat_fullname}
\bibliography{main}
}

\end{document}